\documentclass{article} 
\usepackage{iclr2023_conference,times}

\usepackage{amsmath,amsfonts,bm}

\def\eqref#1{equation~\ref{#1}}

\def\1{\bm{1}}

\def\eps{{\epsilon}}

\DeclareMathAlphabet{\mathsfit}{\encodingdefault}{\sfdefault}{m}{sl}
\SetMathAlphabet{\mathsfit}{bold}{\encodingdefault}{\sfdefault}{bx}{n}

\usepackage[utf8]{inputenc} 
\usepackage[T1]{fontenc}    
\usepackage{booktabs}       
\usepackage{amsfonts}       
\usepackage{nicefrac}       
\usepackage{xcolor}         

\usepackage{hyperref}

\hypersetup{
    colorlinks=false,
    pdfborder={0 0 0},
}

\usepackage{url}

\usepackage{microtype}
\usepackage[pdftex]{graphicx}

\usepackage{subfig}
\usepackage{booktabs} 

\usepackage{multirow}

\usepackage{enumitem}

\usepackage{wrapfig}

\usepackage{collcell}
\usepackage{xstring}
\usepackage{datatool}
\usepackage{booktabs}
\usepackage{environ}

\newcounter{CurrentRow}
\newcounter{CurrentColumn}
\newtoggle{DoneWithFirstRow}

\newcommand*{\FirstColumn}[1]{%
    \IfEq{\arabic{CurrentColumn}}{0}{%
        \global\togglefalse{DoneWithFirstRow}%
        \setcounter{CurrentRow}{1}
    }{%
        \global\toggletrue{DoneWithFirstRow}%
        \stepcounter{CurrentRow}%
    }%
    \setcounter{CurrentColumn}{0}%
    \NewData{#1}%
}
\newcommand*{\NewData}[1]{%
    \dtlexpandnewvalue%
    \stepcounter{CurrentColumn}%
    \iftoggle{DoneWithFirstRow}{%
        \dtlgetrow{TransposedTabularDB}{\arabic{CurrentColumn}}%
        \dtlappendentrytocurrentrow{\Alph{CurrentRow}}{#1}%
        \dtlrecombine%
    }{%
        \DTLnewrow{TransposedTabularDB}%
        \DTLnewdbentry{TransposedTabularDB}{\Alph{CurrentRow}}{#1}%
    }%
}%

\newcolumntype{F}{>{\collectcell\FirstColumn}c<{\endcollectcell}}
\newcolumntype{C}{>{\collectcell\NewData}{c}<{\endcollectcell}}

\newtoggle{EncounteredDataRow}

\newsavebox{\TempBox}
\DTLnewdb{TransposedTabularDB}

\NewEnviron{Ttabular}[1]{%
    \setcounter{CurrentColumn}{0}%
    \global\togglefalse{EncounteredDataRow}%
    \savebox{\TempBox}{%
        \begin{tabular}{FCCCCCC}
            \BODY%
        \end{tabular}%
    }%
    \begin{tabular}{#1}\toprule%
    \DTLforeach*{TransposedTabularDB}{\Aa=A, \Ba=B, \Ca=C}{%
      \DTLiffirstrow{}{\\\midrule}%
        \Aa & \Ba & \Ca %
    }\\\bottomrule%
    \end{tabular}%
}%

\usepackage[colorinlistoftodos]{todonotes}

\usepackage{twoopt}
\usepackage{tikz}
\usetikzlibrary{fit}
\newcommand\tikzmark[1]{%
  \tikz[remember picture,overlay]\node[inner xsep=0pt] (#1) {};}
\newcommandtwoopt\Textbox[5][2.5cm][2cm]{%
\begin{tikzpicture}[remember picture,overlay]
  \coordinate (aux) at ([xshift=#1]#4);
  \node[inner ysep=3pt,yshift=0.6ex,draw=green,thick,
    fit=(#3) (aux),baseline] 
    (box) {};
  \node[text width=#2,anchor=north east,
    font=\sffamily\footnotesize,align=right] 
    at (box.north east) {#5};
\end{tikzpicture}%
}

\usepackage{xspace}

\newcommand{\hprd}{\textsc{HPRD}\xspace}

\newcommand{\BA}{\textsc{BA}\xspace}
\newcommand{\brain}{\textsc{Brain}\xspace}
\newcommand{\cosmos}{\textsc{Cosmos}\xspace}
\newcommand{\ethe}{\textsc{Ethe}\xspace}
\newcommand{\youtube}{\textsc{Youtube}\xspace}
\newcommand{\youtubeS}{\textsc{Yt-S}\xspace}
\newcommand{\youtubeL}{\textsc{Yt-L}\xspace}
\newcommand{\dblp}{\textsc{Dblp}\xspace}
\newcommand{\dblpS}{\textsc{Dblp-S}\xspace}
\newcommand{\dblpL}{\textsc{Dblp-L}\xspace}

\newcommand{\task}{Subgraph Matching\xspace}
\newcommand{\sgmrlmodel}{\textsc{NSubS}\xspace}

\newcommand{\encoderlayer}{\textsc{Qc-Sgmnn}\xspace}

\newcommand{\mcsrlmodel}{\textsc{GLSearch}\xspace}

\newcommand{\ldf}{\textsc{Ldf}\xspace}
\newcommand{\nlf}{\textsc{Nlf}\xspace}
\newcommand{\tso}{\textsc{Tso}\xspace}
\newcommand{\cfl}{\textsc{Cfl}\xspace}
\newcommand{\ceci}{\textsc{Ceci}\xspace}
\newcommand{\qsi}{\textsc{QuickSI}\xspace}

\newcommand{\daf}{\textsc{DP-iso}\xspace}
\newcommand{\gql}{\textsc{GraphQL}\xspace}

\newcommand{\nmatch}{\textsc{NMatch}\xspace}

\newcommand{\valuee}{$v_\theta(s_t)$\xspace}
\newcommand{\policye}{$P_\theta(a_t|s_t)$\xspace}

\newcommand{\numcpp}{seven\xspace}
\newcommand{\numdata}{six\xspace}

\usepackage{amsmath}
\usepackage{amssymb}
\usepackage{amsthm}
\usepackage{bm}
\usepackage{algorithm}
\usepackage{algorithmic}

\usepackage{bbm}
\newcommand{\inv}{^{\raisebox{.2ex}{$\scriptscriptstyle-1$}}}

\title{Detecting Small Query Graphs in A Large Graph via Neural Subgraph Search}

\author{%
  Derek Xu\thanks{Equal contribution.} \\
  University of California Los Angeles
 \\
  \texttt{derekqxu@g.ucla.edu} \\
  \And
  Yunsheng Bai\footnotemark[1] \\
  University of California Los Angeles
 \\
  \texttt{yba@g.ucla.edu} \\
  \AND
  Yizhou Sun \\
  University of California Los Angeles
 \\
  \texttt{yzsun@cs.ucla.edu} \\
  \And
  Wei Wang \\
  University of California Los Angeles
 \\
  \texttt{weiwang@cs.ucla.edu} \\
}

\iclrfinalcopy 
\begin{document}

\maketitle

\begin{abstract}


Recent advances have shown the success of using reinforcement learning and search to solve NP-hard graph-related tasks, such as Traveling Salesman Optimization, Graph Edit Distance computation, etc. However, it remains unclear how one can efficiently and accurately detect the occurrences of a small query graph in a large target graph, which is a core operation in graph database search, biomedical analysis, social group finding, etc. This task is called \task which essentially performs subgraph isomorphism check between a query graph and a large target graph. 
One promising approach to this classical problem is the ``learning-to-search'' paradigm, where a reinforcement learning (RL) agent is designed with a learned policy to guide a search algorithm to quickly find the solution without any solved instances for supervision. However, for the specific task of \task, though the query graph is usually small given by the user as input,  the target graph is often orders-of-magnitude larger. 
It poses challenges to the neural network design 
and can lead to solution 
and reward 
sparsity. In this paper, we propose \sgmrlmodel with two innovations to tackle the challenges:
(1) A novel encoder-decoder neural network architecture to dynamically compute the matching information between the query and the target graphs at each search state; (2) A novel look-ahead loss function for training the policy network.
Experiments on \numdata large real-world target graphs show that \sgmrlmodel can significantly improve the subgraph matching performance. 
\end{abstract}

\section{Introduction}
\label{sec-intro}

With the growing amount of graph data that naturally arises in many domains, solving graph-related tasks via machine learning has gained increasing attention. Many NP hard tasks, 
 e.g. Traveling Salesman Optimization~\citep{xing2020graph}, Graph Edit Distance computation~\citep{wang2021bi}, Maximum Common Subgraph detection~\citep{bai2021glsearch}, have recently been tackled via learning-based methods. These works on the one hand rely on search to enumerate the large solution space, and on the other hand use reinforcement learning (RL) to learn a good search policy from training data, thus obviating the need for hand-crafted heuristics adopted by traditional solvers. Such learning-to-search paradigm~\citep{bai2021glsearch} also allows the training the RL agent without any solved instances for supervision. However, how to design a neural network architecture 
 under the RL-guided search framework remains unclear for the task of \task, which requires the detection of all occurrences of a small query graph in an orders-of-magnitude larger target graph. \task has wide applications in graph database search~\citep{lee2012depth}, knowledge graph query~\citep{kim2015taming}, biomedical analysis~\citep{zhang2009gaddi}, social group finding~\citep{ma2018comparative}, quantum circuit design~\citep{jiang2021quantum}, etc. As a concrete example, \task is used for protein complex search in a protein-protein interaction network to test whether the interactions within a protein complex in a species are also present in other species~\citep{bonnici2013subgraph}.

Due to its NP-hard nature, the state-of-the-art \task algorithms rely on backtracking search with various techniques proposed to reduce the large search space~\citep{sun2020memory,kim2021versatile,wang2022reinforcement}. However, these techniques are mostly driven by heuristics, and as a result, we observe that such solvers
often fail to find any solution on large target graphs under a reasonable time limit,
although they tend to work well on small graph pairs. 
We denote this phenomenon as {\it solution sparsity}. Such solution sparsity requires the designed model to not only have enough capacity but also to run efficiently under limited computational budget. Another consequence of solution sparsity is that, there can be little-to-no reward signals for the RL agent under an RL training framework~\citep{silver2017mastering}, which we denote as {\it reward sparsity}. 



In this paper, we propose \sgmrlmodel with two means to address the aforementioned challenges. First, we propose a novel graph encoder-decoder neural network to dynamically match the query graph with the target graph and perform aggregation operation only on the query graph to reduce information loss. The novel encoder decouples the intra-graph message passing module (the ``propagation'' module) that yields state-independent node embeddings, and the inter-graph message passing module (the ``matching'' module) that refines the node embeddings via subgraph-to-graph matching. Thus, the intra-graph embeddings can be computed only once at the beginning of search for efficient inference. 
We further advance the inter-graph message passing by propagating only between nodes that either are already matched or can be matched in future by running a local candidate search space computation algorithm at each search state. Such algorithm leverages the key requirement of \task that every node and edge in the query graph must be matched to the target graph, and therefore reduces the amount of candidates from all the nodes in the target graph to a much smaller amount. Compared with a Graph Matching Network~\citep{li2019graph} which computes all the pairwise node-to-node message passing between two input graphs, our matching module is able to focus on only the node pairs that can contribute to the solution, and thus is both more effective and more efficient.
In addition, we propose the use of sampling of subgraphs to obtain ground-truth subgraph-to-graph node-node mappings to alleviate the reward sparsity issue during training. We design a novel look-ahead loss function where the positive node-node pairs are augmented with positive node-node pairs in future states to boost the amount of training signals at each search state. 

Experiments on synthetic and real graph datasets demonstrate that \sgmrlmodel outperforms baseline solvers in terms of effectiveness by a large margin. Our contributions can be summarized as follows:
\begin{itemize}
\item We address the challenging yet important task of \task with a vast amount of practical applications and propose \sgmrlmodel as the solution.
\item One key novelty is a proposed encoder layer consisting of a propagation module and a matching module that dynamically passes the information between the input graphs. 
\item We conduct extensive experiments on real-world graphs to demonstrate the effectiveness of the proposed approach compared against a series of strong baselines in \task.
\end{itemize}


\section{Preliminaries}
\label{sec-prelim}

\subsection{Problem Definition}
\label{subsec-probdef}

We denote a query graph as $q=(V_q,E_q)$ and a target graph as $G=(V_G,E_G)$ where $V$ and $E$ denote the node and edge sets. $q$ and $G$ are associated with a node labeling function $L_g$ which maps every node into a label $l$ in a label set $\Sigma$. \enspace  \textbf{Subgraph}: For a subset of nodes $S$ of $V_q$, 
$q[S]$ denotes the subgraph of $q$ with an node set $S$ and a edge set consisting of all the edges in $E_q$ that have both endpoints in $S$. 
In this paper, we adopt the definition of non-induced subgraph.
\enspace \textbf{Subgraph isomorphism}: $q$ is subgraph isomorphic to $G$ if there exists an injective node-to-node mapping $M: V_q \rightarrow V_G$ such that (1) $\forall u \in V_q,L_g(u)=L_g(M(u))$; and (2) $\forall e_{(u,u')} \in E_q,e_{(M(u),M(u'))} \in E_G$.  
\enspace \textbf{\task}: The task of \task aims to find the subgraphs in $G$ that are isomorphic to $q$. We call $M$ a solution, or a match of $q$ to $G$. We call a pair $(q,G)$ is solved if the algorithm can find any match under a given time limit, which we find a challenge for existing solvers on input graphs in experiments especially on large graphs.
For solved pairs, the number of found subgraphs by an algorithm is reported.

\subsection{Related Work} 
\label{sec-related}

\paragraph{Non-learning methods on \task}

Existing methods on \task can be broadly categorized into backtracking search algorithms~\citep{shang2008taming,he2008graphs,han2013turboiso,han2019efficient,kim2021versatile,wang2022reinforcement} and multi-way join approaches~\citep{lai2015scalable,lai2016scalable,lai2019distributed,kankanamge2017graphflow}. 
The former category of approaches employ a branch and bound approach to grow the solution from an empty subgraph by gradually seeking  one matching node pair at a time following a strategic order until the entire search space is explored. 
The multi-way join approaches rely on decomposing the query graph into nodes and edges and performing join operations repeatedly to combine the partially matched subgraphs to $q$. However, they tend to work well on small query graphs generally with less than 10 nodes~\citep{sun2020memory}, and thus we follow and compare against methods in the former category, whose details will be shown in Section~\ref{subsec-search}. 

\paragraph{Learning-based methods for subgraph-related problems} 
The idea of designing neural networks for predicting subgraph-graph relation has been explored, which due to its approximate nature may contain false positives/negatives, and it remains unclear how one can directly use these methods for exact \task. For example, \nmatch~\citep{lou2020neural} learns node embeddings to predict a score for an input subgraph-graph pair indicating whether the subgraph is contained in another graph, which can be regarded as solving the approximate version of \task.
Another direction of research aims to perform subgraph counting~\citep{liu2020neural,chen2020can} supervised on the number of specific substructures, and again lacks an explicitly search strategy and thus falls short of yielding solutions for \task.

\paragraph{Efforts on using RL for graph NP-hard problems} The idea of using RL to replace heuristics in search algorithms for NP-hard graph-related tasks is not new, and we identify three works similar to the present work. (1) \textsc{GLSearch}~\citep{bai2021glsearch} detects the maximum common subgraph (MCS) in an input graph pair, which is different from \task which
requires the entire $q$ to be matched with $G$, allowing further improvement in the neural network and search design. (2) \textsc{RL-QVO}~\citep{wang2022reinforcement} tackles \task via ordering the nodes in the \textit{query} graph as a global pre-processing step before search, which is an orthogonal direction to our approach to select nodes in the \text{target} graph computed at each search step.

\subsection{Search-based methods for \task}
\label{subsec-search}



  \begin{wrapfigure}{R}{0.5\textwidth}
    \begin{minipage}{0.5\textwidth}

\begin{algorithm}[H]
	\caption{Search-based \task.
	}
	\begin{algorithmic}[1]
		\STATE \textbf{Input:} Query graph $q$, data graph $G$.
		
		\STATE \textbf{Output:} Matches from $q$ to $G$.

		\STATE Filter: $C \leftarrow$ generate candidate node sets.
		
		\STATE Order: $\phi \leftarrow$ generate an ordering for $V_q$.
		
		\STATE Search: $\texttt{Backtracking}(q,G,C,\phi,\{\})$.

	\end{algorithmic}
	\label{algo-overall}

\end{algorithm}

    \end{minipage}
  \end{wrapfigure}


Due to the NP-hard nature of \task, backtracking search is a naturally suitable algorithm since it exhaustively explores the solution space by starting with an empty match and adding one new node pair to the current match at each step. When the current match cannot be further extended, the search backtracks to its previous search state, and explores other node pairs to extend the match. However, naively enumerating all the possible states in the entire search space is intractable in practice, and therefore existing efforts mainly aim to reduce the total number of search steps 



  \begin{wrapfigure}{R}{0.5\textwidth}
    \begin{minipage}{0.5\textwidth}

\begin{algorithm}[H]
	\caption{ $\texttt{Backtracking}(q,G,C,\phi,M)$
	}
	\begin{algorithmic}[1]
		\STATE \textbf{Input:} $q$, $G$, $C$, $\phi$, and current mapping $M$.
		
		\STATE \textbf{Output:} Subgraph match mappings.

        \IF {$|M| = |V_q|$}
            \STATE output $M$;
            \STATE return;
        \ENDIF
        
        \STATE 
        $u_t \in V_q \leftarrow \phi(M)$;

       \STATE 
        $\mathcal{A}_{u_t} \leftarrow s_{t}.getLocalCand(u_t)$;
        
        \STATE
        \tikzmark{start4}  $\mathcal{A}_{u_t,\mathrm{ordered}} \leftarrow policy(s_{t}, \mathcal{A}_{u_t})$;   \tikzmark{end4}
    
        \FOR {$v_t$ in  $\mathcal{A}_{u_t,\mathrm{ordered}}$}

            
            \STATE $M \leftarrow M.add(u_t,v_t)$;
            
            \STATE
            $\texttt{Backtracking}(q,G,C,\phi,M)$;
            
          \STATE $M \leftarrow M.remove(u_t,v_t)$;
        
        \ENDFOR
        
	\end{algorithmic}
	\label{algo-backtrack}

\end{algorithm}

    \end{minipage}
  \end{wrapfigure}

for the backtracking search via mainly three ways~\citep{sun2020memory}: (1) Filter nodes in $G$ to obtain a small set of candidate nodes for each node in $q$ as a pre-processing step before the backtracking search; (2) Order the nodes in $q$ before the search; (3) Generate a local candidate set of nodes in each step of the search based on the current search state.
Algorithm~\ref{algo-overall} summarizes the overall backtracking search based framework for \task. It is worth noting that the first three means correspond to the three steps in the algorithm, and therefore any improvement in any of the three steps can be regarded as orthogonal to each other. 

The basic idea of backtracking search is outlined in Algorithm~\ref{algo-backtrack}. The recursive algorithm starts with an empty node-node mapping, and tries to add one new node pair to the mapping $M$ at each recursive call.
The action is the new node pair $(u_t,v_t)$, where $u_t \in V_q$ is selected according to the heuristic-based ordering $\phi$,
and $v_t \in V_G$ is selected according to a policy (which is to be learned by \sgmrlmodel) to be one of the local candidate nodes (line 8), 
that can be mapped to $u_t$. It is noteworthy that this local candidate node set ``$\mathcal{A}_{u_t}\subseteq V_G$'' 
is refined over the global candidate sets $C$ based on the current search state $s_t$. $s_t$ is defined as $(q,G)$ along with the current mapping $M$
and $u_t$.
$\mathcal{A}_{u_t}\subseteq V_G$ ensures any node in $\mathcal{A}_{u_t}$ would lead to the extended subgraphs at $s_t$ 
still being isomorphic to each other. 
Thus, the local candidate set $\mathcal{A}_{u_t}$ for $u_t$ is the action space, for which we learn a policy  (line 9) to order the nodes, resulting in an ordered list $\mathcal{A}_{u_t,\mathrm{ordered}}$. 

Despite existing efforts to compute a small $C$, a good $\phi$, and a small $\mathcal{A}_{u_t}$ to reduce search space, we observe that the size of $\mathcal{A}_{u_t}$ can be up to thousands of nodes for many real-world large target graphs, calling for a smarter policy to order the nodes not only in $q$ but also in $G$ ($\mathcal{A}_{u_t}$ to be specific). To the best of our knowledge, all the existing methods adopt a random ordering in $\mathcal{A}_{u_t,\mathrm{ordered}}$. 
We conjecture it is because existing local candidate computation techniques can further prune nodes from $C$, and thus enumerating $\mathcal{A}_{u_t}$ in a random ordering can be tractable on small graph pairs. We will experimentally show that this attributes to their failure to find a match for many large graph pairs. In fact, a theoretically perfect policy, $policy^{*}$, can find the entire match of the query graph in $V_q$ steps or recursive calls, assuming $q$ has at least one match with $G$. This again inspires our proposed method to improve the policy for node selection from $\mathcal{A}_{u_t}$. 

\section{Proposed Method}
\label{sec-model}

In this section we formulate the problem of \task as learning an RL agent that grows the extracted subgraphs by adding new node pairs to the current subgraphs. 
We first describe the environment setup, then depict our proposed encoder-decoder neural network which provides actions for our agent to grow the subgraphs in a search context.

\subsection{RL Formulation of \task} 
\label{subsec-rlform}

We formulate the \task problem as a Markov Decision Process (MDP), where the RL agent starts with an empty solution, and iteratively matches one node pair at a time until no more nodes can be matched. As mentioned in Section~\ref{subsec-search}, our policy assigns a score to each action in  $\mathcal{A}_{u_t}$, and therefore can be modeled as a policy network \policye that computes a probability distribution over $\mathcal{A}_{u_t}$ for the current state $s_t$, where the node pair to select consists of $(u_t,v_t)$. However, since $u_t$ is chosen by $\phi$, we regard an action as a node $v_t$ selected from $\mathcal{A}_{u_t}$. Since each action is a target node in $\mathcal{A}_{u_t}$, the neural network must learn good node embeddings that capture the current matching status in each search state, which will be shown in Section~\ref{subsec-policy}. We define our reward as $R^{(+)}=1$ if the subgraph is fully matched and $R^{(-)}=0$  otherwise. 

To provide ample training signals to the neural network model, during training, we randomly sample subgraphs from the target graph and record the mapping between the nodes in the sampled subgraph and $V_{G}$. Therefore, after the search, we collect the positive training signals at each state as all the node-node pairs $(u,v)$ that lead to a solution. 
The collection of such positive pairs is denoted as $P(s_t)$. Negative pairs $N(s_t)$ can be obtained by randomly sampling node-node pairs at each $s_t$ from the action space $\mathcal{A}_{u_t}$. However, the solution sparsity issue mentioned previously causes insufficient positive pairs. We therefore propose the following look-ahead loss function which augments $P(s_t)$ with the positive node-node pairs in future states. Overall, the look-ahead loss function can be formulated as
\begin{equation}
\label{eq-L1}
L_{\mathrm{la}}(s_{t}) =  \sum_{k=0}^{T-t} L_{\mathrm{ce}}(s_{t+k})
\end{equation} 
where $T$ corresponds to the final step that matches the entire subgraph, and $L_{\mathrm{la}}'$ is the cross-entropy loss computed at $s_{t+k}$. Specifically, $L_{\mathrm{ce}}(s_{t+k})$ is the cross entropy loss and computed as the negative of the following:
\begin{equation}
\label{eq-L1prime}
\begin{array}{rl}
&\sum_{(u, v) \in P({s_{t+k}})}  R^{(+)} \mathrm{log} \pi_\theta \big((u,v)|s_t \big) +  (1-R^{(+)}) \mathrm{log} \Big( 1 - \pi_\theta \big( (u,v)|s_t \big) \Big) + \\
&\sum_{(u, v) \in N({s_{t+k}})}   R^{(-)} \mathrm{log} \pi_\theta \big((u,v)|s_t \big) +  (1-R^{(-)}) \mathrm{log} \Big( 1 - \pi_\theta \big( (u,v)|s_t \big) \Big)
\end{array}
\end{equation}
where $\pi_\theta \big((u,v)|s_t \big) = \sigma \Big( P_{\mathrm{logit}}\big((u,v)|s_t\big) \Big)$ denotes the raw output of the policy network followed by applying the Sigmoid function. Since we define $R^{(+)}=1$ and $R^{(-)}=0$, $L_{\mathrm{ce}}(s_{t+k})$ is simplified as
\begin{equation}
\label{eq-L1prime_simple}
\begin{array}{rl}
- \sum_{(u, v) \in P({s_{t+k}})} \mathrm{log} \pi_\theta \big((u,v)|s_t \big) - \sum_{(u, v) \in N({s_{t+k}})} \mathrm{log} \Big( 1 - \pi_\theta \big( (u,v)|s_t \big) \Big)
\end{array}
\end{equation}


In addition to the look-ahead loss that encourages the policy network to recognize positive node-node pairs, we also adopt the max margin loss, $L_{\mathrm{mm}}$, proposed in \citet{lou2020neural} to encourage the encoder to produce subgraph-graph relation aware node embeddings $\bm{h}$. Our overall loss function is the combination of the novel look-ahead loss function and the max margin loss function, i.e. $L_{\mathrm{total}}= L_{\mathrm{la}} + L_{\mathrm{mm}}$. More details can be found in the appendix.

\subsection{Encoder-Decoder Design for Policy Estimation}
\label{subsec-policy}

Our neural network consists of an encoder which produces node-level embeddings for $V_q$ and $V_G$
and a decoder which transforms the node embeddings into \policye.
We identify the following challenges: (1) Each state $s_t$ consists of $q$, $G$, 
and the mapping $M_t$ between the matched nodes $S_{t,q}\subseteq V_q$ and $S_{t,G}\subseteq V_G$, which should be effectively represented and utilized in predicting \policye; (2) \policye is dependent on $s_t$ and $a_t$, potentially requiring the node embeddings to be recomputed at every search step and incurring too much computational overhead;
(3) Existing Graph Neural Networks (GNN) relying on message passing (e.g. \citet{kipf2016semi,velickovic2017graph,you2020design}) are inherently local, which are incompatible with the policy estimation requiring global matching of $q$ to $G$. Even the more recent GNNs that enhance expressiveness beyond 1-Weisfeiler-Lehman test~\citep{morris2020weisfeiler,wijesinghe2021new} still cannot guarantee the capturing of nodes in the solution, since otherwise the NP-hard task would then be solved. 
This motivates the following design leveraging the properties of \task as much as possible.

\subsubsection{Encoder: Embedding Generation via \encoderlayer}
\label{subsec-encoder}

\begin{figure*}
\centering
\includegraphics[width=0.99\textwidth]{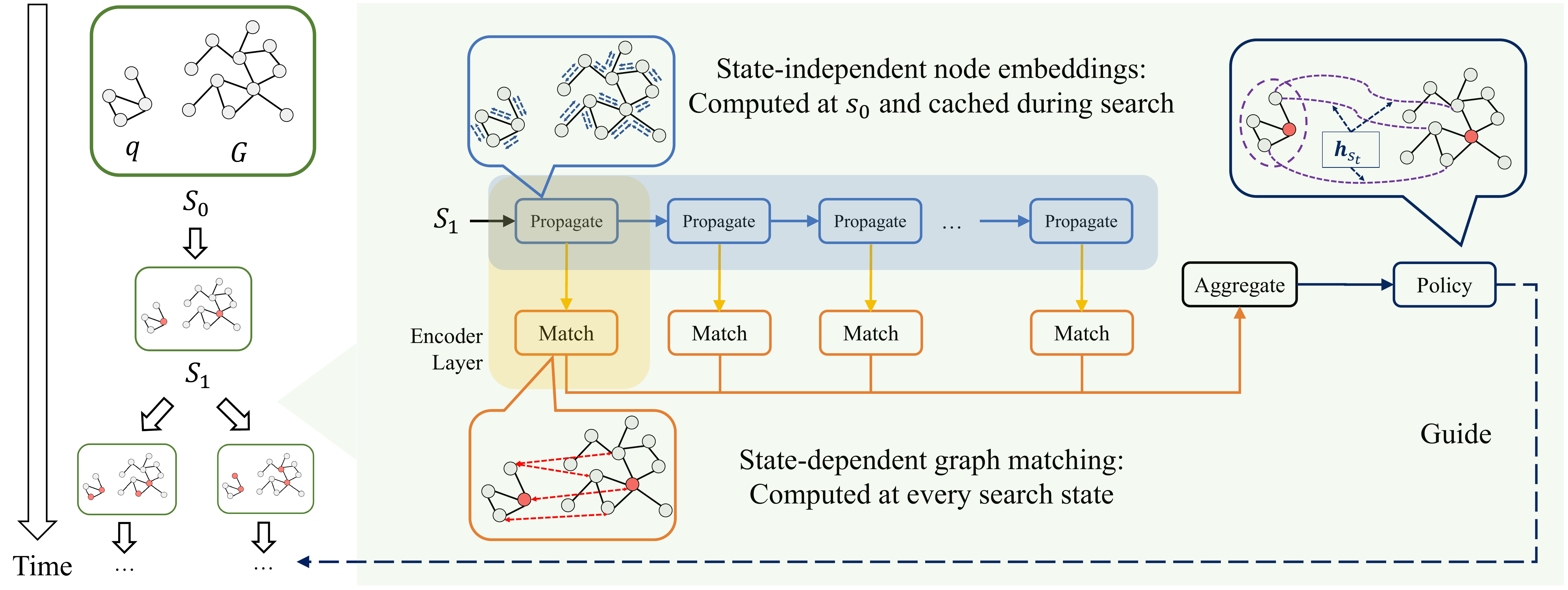}
\caption{The overall process of \task is a search algorithm that matches one node pair at a time for the input query $q$ and target graph $G$ guided by a learned policy. 
Due to the large action space incurred by the large $G$ in practice, we propose to train a policy network to guide the selection of local candidate nodes in $G$ at each search state. This requires effective node embeddings to be learned that can reflect the node-to-node mapping at the current state and contribute to the prediction of the policy and reward. To achieve this end, we propose a novel \encoderlayer encoder layer consisting of a ``Propagation'' module that performs intra-graph message passing as a typical GNN and a ``Matching'' module that performs state-dependent inter-graph matching (Section~\ref{subsec-encoder}). The \policye is trained using the novel loss functions described in Section~\ref{subsec-rlform}.
}
\label{fig:model}
\vspace*{-4mm}
\end{figure*}

The nodes of $q$ and $G$ are encoded into initial one-hot embeddings $\bm{h}^{(0)} \in \mathbb{R}^{|\Sigma|}$, and fed into $K$ sequentially stacked Query-Conditioned Subgraph Matching Neural Networks (\encoderlayer) layers to obtain $\bm{h}^{(K)} \in \mathbb{R}^{D}$, where $D$ is the dimension of embeddings, followed by a \textsc{Jumping Knowledge}~\citep{xu2018representation} network to combine the node embeddings from the $K$ layers to obtain the final node embeddings, denoted as ``Aggregate'' in Figure~\ref{fig:model}.

We start by observing that \task desires all nodes in $q$ to be matched with $G$, and existing works in \task aiming at reduce $\mathcal{A}_{u_t}$ (Section~\ref{subsec-search}) guarantee no false pruning. In other words, $\mathcal{A}_{u_t}$ is guaranteed to contain the solution nodes in $G$ that should match with $u_t$, that is, $\mathcal{A}_{u_t}$ provides useful hints of which nodes should be matched together.
We then compute the candidate space $\mathcal{A}_{u'}$ for every remaining query graph node $u' \notin S_{t,q}$, resulting in a mapping
$M_t' = \{ u' \mapsto \mathcal{A}_{u'}, \forall u' \in V_q \setminus S_{t,q}  \} $. 
Intuitively, $M_t'$ can be regarded as hints of \textit{future} node-node mappings. In contrast, $M_t = \{ u \mapsto \{ v \}, \forall u \in S_{t,q} \} $ reflects the \textit{current} state $s_t$, which maps each node in $S_{t,q}$ to a unique node in $G$. We define $\tilde{M_t}$ as the union of $M_t$ and $M_t'$, and define  $\tilde{M_t\inv}$ as the reverse mapping of $\tilde{M_t}$, which serve as the basis for the matching module.

\encoderlayer consists of a propagation module, which performs regular intra-graph message passing on $q$ and $G$ individually, and a matching module, which performs state-dependent subgraph-graph matching leveraging $\tilde{M_t}$ to pass 
information between $q$ and $G$. Specifically, 
\begin{equation}
\label{eq-encoderlayer}
\begin{array}{rl}
\bm{h}_{u,\mathrm{intra}}^{(k+1)} =& f_{\mathrm{agg}} \big(\{ f_{\mathrm{msg}} (\bm{h}_{u,\mathrm{intra}}^{(k)}, \bm{h}_{u',\mathrm{intra}}^{(k)})  | u' \in \mathcal{N}(u) \} \big), \\

\bm{h}_{v,\mathrm{intra}}^{(k+1)} =& f_{\mathrm{agg}} \big(\{ f_{\mathrm{msg}} (\bm{h}_{v,\mathrm{intra}}^{(k)}, \bm{h}_{v',\mathrm{intra}}^{(k)})  | v' \in \mathcal{N}(v) \} \big), \\

\bm{h}_{q,\mathrm{intra}}^{(k+1)} =& f_{\mathrm{readout}} ( \{ \bm{h}_{u,\mathrm{intra}}^{(k+1)}  | u \in V_q \} ), \\

\bm{h}_{u,G \rightarrow q}^{(k+1)} =& f_{\mathrm{agg}} \big(
\{ f_{\mathrm{msg}} (\bm{h}_{u,\mathrm{intra}}^{(k+1)}, \bm{h}_{v,\mathrm{intra}}^{(k+1)})  | v \in \tilde{M_t}(u) \} \big), \\

\bm{h}_{v,q \rightarrow G}^{(k+1)} =& f_{\mathrm{agg}} \big(
\{ f_{\mathrm{msg}} (\bm{h}_{v,\mathrm{intra}}^{(k+1)}, \bm{h}_{u,\mathrm{intra}}^{(k+1)}, \bm{h}_{q,\mathrm{intra}}^{(k+1)})  | u \in \tilde{M_t\inv}(v) \} \big), \\

\bm{h}_{u}^{(k+1)} =& f_{\mathrm{combine}} ( \bm{h}_{u,G \rightarrow q}^{(k+1)},
\bm{h}_{u,\mathrm{intra}}^{(k+1)} ), \\

\bm{h}_{v}^{(k+1)} =&
f_{\mathrm{combine}} (
\bm{h}_{v,q \rightarrow G}^{(k+1)},
\bm{h}_{v,\mathrm{intra}}^{(k+1)}).
\end{array}
\end{equation}
The first two steps can be any intra-graph message passing GNNs such as Graph Attention Networks~\citep{velickovic2017graph} with a message function $f_{\mathrm{msg}}$ and an aggregation function $f_{\mathrm{agg}}$, corresponding to the propagation module.  The middle three steps compute intermediate embeddings  that will be used for the last two steps, i.e. the matching module. Specifically, $\bm{h}_{u,G \rightarrow q}^{(k+1)}$ and $\bm{h}_{v,q \rightarrow G}^{(k+1)}$ compute the cross-graph message passing from $G$ to $q$ and $q$ to $G$ using $\tilde{M_t}$ and $\tilde{M_t\inv}$, respectively (represented as the red dashed lines in Figure~\ref{fig:model}). A graph-level embedding $\bm{h}_{q,\mathrm{intra}}^{(k)}$ is computed via $f_{\mathrm{readout}}$ and used in the information passing from $q$ to $G$ to let the embeddings of $V_G$ query-conditioned. We do not inject the graph-level embeddings of $G$ into $\bm{h}_{u,G \rightarrow q}^{(k+1)}$, since the large size of $G$ could result in too much information loss in the readout operation. The last two steps combine the intra-graph embeddings and inter-graph embeddings via $f_{\mathrm{combine}}$ to produce the final output embeddings. It is noteworthy that $f_{\mathrm{msg}}$ and $f_{\mathrm{agg}}$ refer to the general class of functions that yield messages between two nodes and performs aggregation on a set of messages, and in practice, the propagation and matching modules can use different functions for $f_{\mathrm{msg}}$ and $f_{\mathrm{agg}}$. 

To address challenge (2), we make the observation that $\bm{h}_{\mathrm{intra}}^{(k)}$ only depends on $q$ and $G$ and is independent of $M_t$, and thus can be computed once and cached at the beginning of the search (denoted as the top branch in Figure~\ref{fig:model}) and later reused throughout the search. 
Our \encoderlayer decouples the propagation and matching steps and outputs two sources of information separately, allowing the search to cache the state-independent node embeddings and dynamically select the computational paths during search.
Thanks to the caching, only the initial iteration requires the $\mathcal{O}(|E_q|+|E_G|)$ computation, and all the subsequent iterations only involve $\mathcal{O}\big(|V_q||\bar{\mathcal{A}}_{u_t}|\big)$ complexity, where $|\bar{\mathcal{A}}_{u_t}|$ is the average size of local candidate space.


\subsubsection{Decoder: \policye Estimation}

Since the node embeddings of $q$ has received the right amount of information from $\tilde{M_t}$, we propose an attention-based mechanism to compute the state embedding: $\bm{h}_{s_t}
= \sum_{u \in V_q} f_{\mathrm{att}} (\bm{h}_{u}, \{ \bm{h}_u' | u' \in V_q \}) \bm{h}_{u}$,
where $f_{\mathrm{att}}$ computes one attention score per node normalized across $V_q$ to tackle challenge (3). Intuitively, the attention function learns which nodes are important for contributing to the eventual subgraph selected by $s_T$. 
Due to the cross-graph communication in \encoderlayer, we only aggregate nodes from $V_q$ to obtain the state representation $\bm{h}_{s_t}$, taking advantage of the fact that $|V_q|$ is typically much smaller than $|V_G|$ in \task, further addressing challenge (2).

For the policy, we aim to tackle challenge (3), by using $\bm{h}_{s_t}$. The reasons are two-fold. First, by definition \policye requires $s_t$ as input; Second, the attention mechanism used to compute $\bm{h}_{s_t}$ capturing the future subgraph. 
Combined with a bilinear tensor product with learnable parameter $\bm{W}^{[1:F]} \in \mathbb{R}^{D \times D \times F}$ with a hyperparameter $F$ to allow the action node embeddings $\bm{h}_{u_t}$ and $\bm{h}_{v_t}$ to fully interact, we obtain $P_{\mathrm{logit}}(a_t|s_t)=\mathrm{MLP} \big( \mathrm{CONCAT} (\bm{h}_{u_t}^{T} \bm{W}^{[1:F]} \bm{h}_{v_t}, \bm{h}_{s_t} ) \big)$,
followed by a $\mathrm{softmax}$ normalization over the logits for all the actions. The decoder has time complexity $\mathcal{O}\big( |V_q| + |\bar{\mathcal{A}}_{u_t}|  \big)$.

\section{Experiments}
\label{sec-exp}


We evaluate \sgmrlmodel against \numcpp backtracking-based algorithms
for exact \task, and conduct experiments on \numdata real-word target graphs from various domains, whose details can be found in the supplementary material. We find \sgmrlmodel written in Python can outperform the state-of-the-art solver under most cases, suggesting the effectiveness of the learned policy for ordering nodes in $G$. Code, trained model, and all the datasets used in the experiments are released as part of the supplementary material for reproducibility. More results and the ablation study can be found in the supplementary material.


\subsection{Datasets and Evaluation Protocol}
\label{subsec-data}

\begin{wraptable}{r}{6.8cm}
\footnotesize
\caption{Target graph description.}\label{table-datasets}
    \begin{tabular}{c|c|c|c}
    \textbf{Dataset} & \textbf{Domain} & \textbf{$|V_G|$} & \textbf{$|E_G|$} \\
    \hline
    \textbf{\BA} & Synthetic & 10,000 & 29,991 \\ \hline
    \textbf{\brain} & Biology & 1,076 & 90,811 \\ \hline
    \textbf{\hprd} & Biology & 9,045  & 34,853 \\ \hline
    \textbf{\cosmos} & Astronomy & 21,596 & 176,830 \\ \hline
    \textbf{\dblp} & Citation & 317,080 & 1,049,866 \\ \hline
    \textbf{\ethe} & Financial & 709,351 & 946,582 \\ \hline
    \textbf{\youtube} & Social & 1,134,890 & 2,987,624 \\
\end{tabular}
\end{wraptable} 
We use one synthetic dataset \BA and \numdata real-world target graphs. For \dblp and \youtube, we prepare two query sets of small and large query graphs, denoted as ``-S'' and ``-L'', respectively, i.e. \dblpS, \dblpL, and \youtubeS, \youtubeL. As shown in Table~\ref{table-datasets}, the largest target graph, \youtube, has over 1M nodes and 2M edges.

To train our model, we sample query graphs of any sizes and let \sgmrlmodel perform the search algorithm and collect training data from the search tree. Therefore, we sample query graphs from each target graph of sizes ranging from 4 nodes 
to 128 nodes, and validate that no query graph in the testing set is visible to the model, that is, no graph in the training set is isomorphic to any graphs in the testing set.
During the testing stage, for each target graph, 
we use the following evaluation protocol. For each graph pair, we set a time limit of 5 minutes, and record the result of whether a solution is found or not, and the number of solutions for solved pairs.

\subsection{Baselines and Parameter Settings}
\label{subsec-baselines}
We compare \sgmrlmodel against a series of baseline solvers whose source codes are provided by \citep{sun2020memory}:
\ldf~\citep{sun2020memory},
\nlf~\citep{sun2020memory}, \qsi~\citep{shang2008taming}, \gql~\citep{he2008graphs},
\tso~\citep{han2013turboiso},
\cfl~\citep{bi2016efficient}, \ceci~\citep{bhattarai2019ceci}, and \daf~\citep{han2019efficient}.
We also include \nmatch~\citep{lou2020neural} as a learning-based method for exact \task, by adapting the original implementation provided by the authors of \citep{lou2020neural} into our backtracking search framework. Specifically, at each search step, we invoke the inference step of the released trained \nmatch model over the input graphs, and obtain a matching score for each $(u_t,v_t)$ action node pair to guide the selection of nodes in $\mathcal{A}_{u_t}$.

Since we are among the first to learn a policy for selecting nodes in $G$, we use \daf for filtering and \gql~ for query node ordering. 
We implement 
the backtracking search framework and \daf's local candidate computation algorithm in Python.

\sgmrlmodel uses 8 layers of the proposed \encoderlayer encoders. $\omega$ is set to 10 and $\mathrm{max\_ iters}$ is set to 120.  Training is performed on a server with NVIDIA Tesla V100 GPUs. We train \sgmrlmodel for approximately 2 days with randomly sampled subgraphs for each target graph with the AdamW optimizer~\citep{loshchilov2017decoupled} with the initial learning rate 0.0005 and the $\eps$ parameter set to 0.01 for numerical stability. Gradient clipping is applied with the maximum norm of the gradients set to 0.1. We use a validation set to select the best model to evaluate.

\begin{figure}[h]
\vspace*{-4mm}
     \centering
     \subfloat[\textbf{\BA}]{\includegraphics[width=0.33\textwidth]{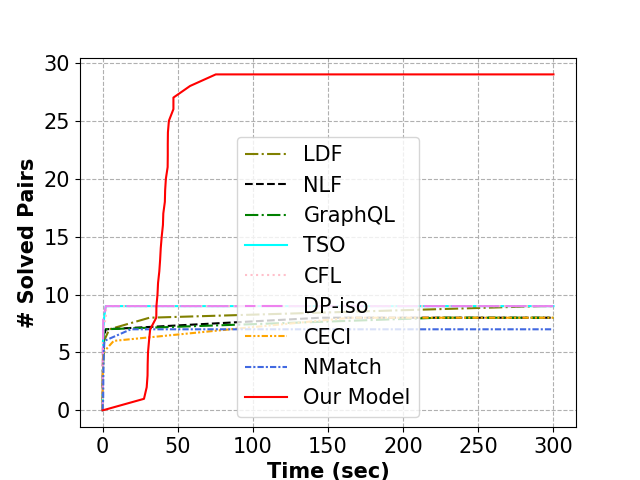}}
     \subfloat[\textbf{\brain}]{\includegraphics[width=0.33\textwidth]{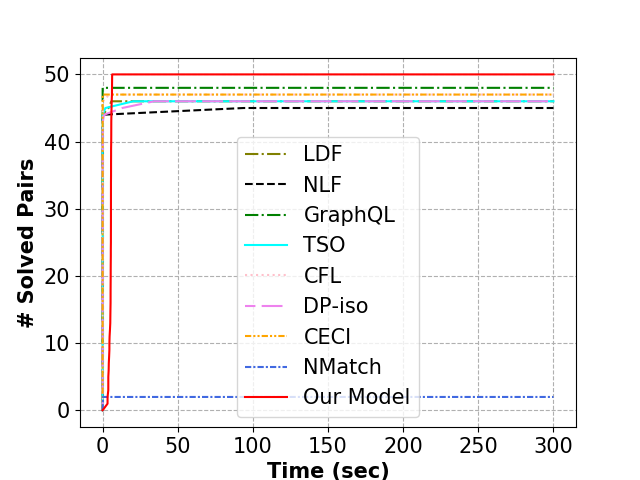}}
     \subfloat[\textbf{\hprd}]{\includegraphics[width=0.33\textwidth]{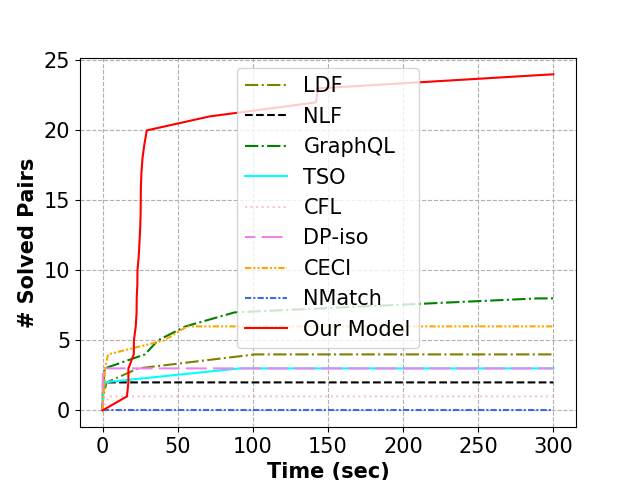}} \\
    \subfloat[\textbf{\cosmos}]{\includegraphics[width=0.33\textwidth]{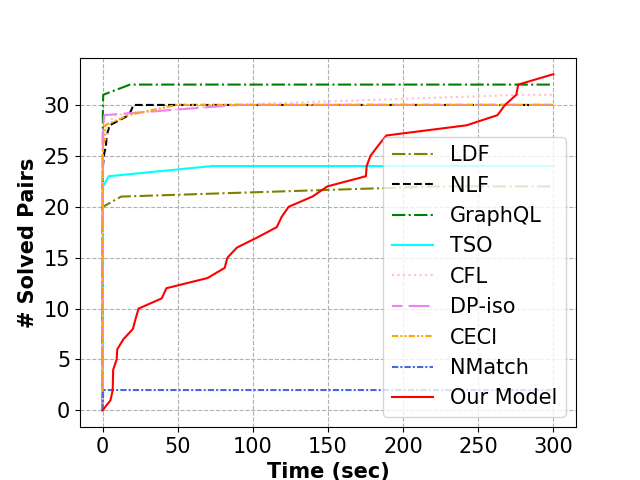}}
     \subfloat[\textbf{\dblpS}]{\includegraphics[width=0.33\textwidth]{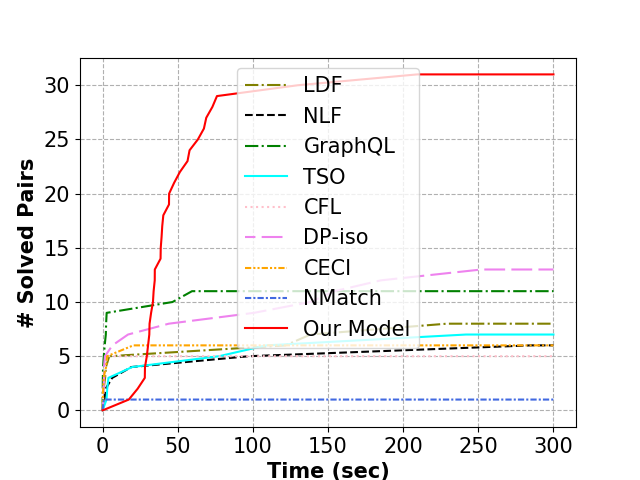}}
     \subfloat[\textbf{\dblpL}]{\includegraphics[width=0.33\textwidth]{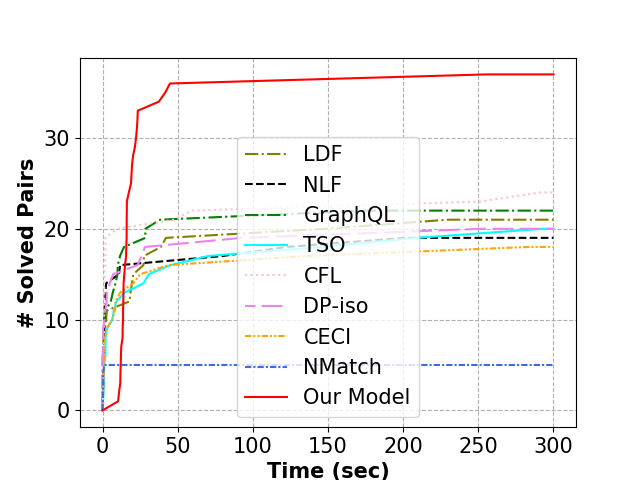}} \\
     \subfloat[\textbf{\ethe}]{\includegraphics[width=0.33\textwidth]{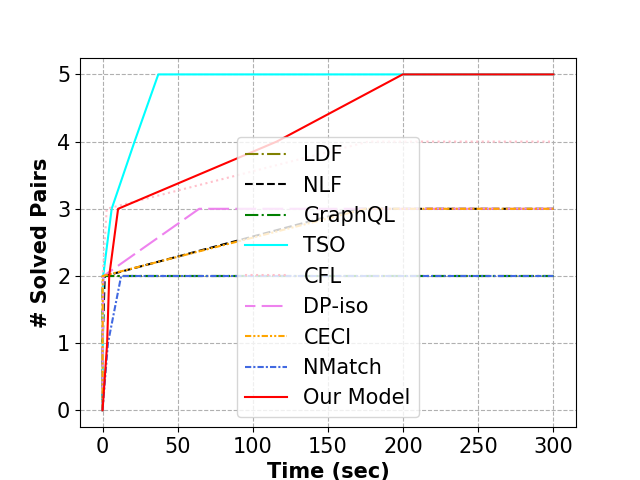}}
     \subfloat[\textbf{\youtubeS}]{\includegraphics[width=0.33\textwidth]{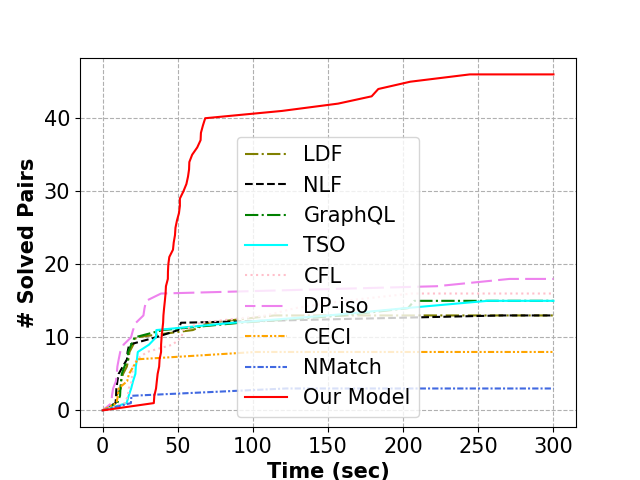}}
     \subfloat[\textbf{\youtubeL}]{\includegraphics[width=0.33\textwidth]{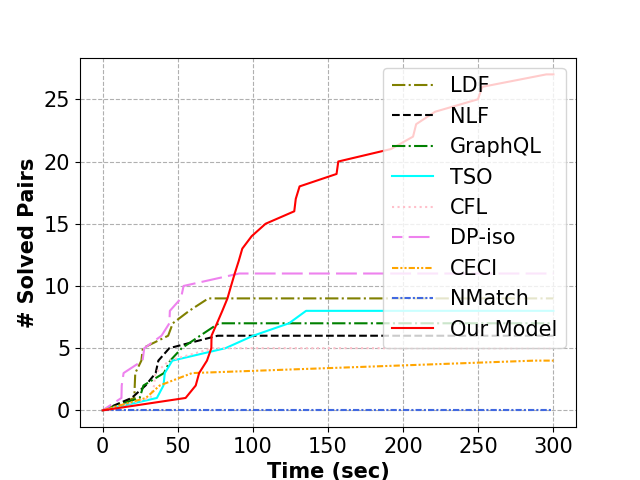}}
     \caption{The growth of the number of solved pairs across time.
     }
     \label{fig:result-curves}
\end{figure}

\subsection{Results}
\label{subsec-result}

To examine the efficiency of each method and analyze the efficacy across time, we conduct the following evaluation. For each $(q,G)$ pair, we record the time the method takes to find a solution, and accumulate the number of solved pairs across time. From $t=0$ to $t=300$ seconds, an increase at $t$ indicates the method solves one graph pair at $t$. The earlier a method solves the graph pairs, the faster and better the method is. 

In theory, given an infinite amount of time, every method adopting the backtracking search algorithm would be able to solve a graph pair. However, such assumption is not practical. Thus, the curves in Figure~\ref{fig:result-curves} have the practical implication that under a reasonable amount of time budget, the idea of using a smarter policy indeed brings better performance to \task. Another observation is that the baseline solvers flatten towards the end of 5 minutes, indicating that they get stuck in unpromising search states that are unlikely to contain the solution, confirming the severity of the aforementioned challenges of solution and reward sparsity.

As shown in Figure~\ref{fig:result-curves}, \sgmrlmodel  
initially seems not as good as other models in the first few seconds due to the additional neural network computational overhead at each step, but achieves the same or better performance on all of the 9 query sets after 5 minutes. 
We observe baseline solvers tend to fail on larger target graphs, as search space pruning on its own is not effective enough to guarantee solutions in such cases with a large amount of candidate nodes. Learning-based methods outperform solver baselines because they provide a better target graph node ordering, under the same search framework. 

Out of the learning-based methods, \sgmrlmodel consistently outperforms \nmatch, as it features a more powerful encoder that efficiently computes a policy conditioned on the current search state with a better inter-graph communication mechanism. This also suggests that the target graph node ordering provided by \sgmrlmodel is much better than the random ordering adopted by solver baselines, confirming that our novelties indeed greatly improve \task performance.

We observe that when the query graph size increase, all methods tend to show lower performance, which can be attributed to the exponentially growing search space. It is noteworthy that the survey paper comparing existing solvers~\citep{sun2020memory} uses query graphs up to 32 nodes, whereas we challenge all methods by testing on query graphs up to hundreds of nodes. 
The fact that \sgmrlmodel is able to solve more graph pairs than baselines when the target graphs are large demonstrates good scalability of \sgmrlmodel.

\begin{table}
\footnotesize
  \begin{center}
    \caption{Number of solutions found by each method after 5 minutes averaged by the number of the graph pairs in each dataset. For clarity and compactness, each result has been divided by 1000, i.e. each number is in the unit of $10^3$.}
    \begin{tabular}{l|lllllllll}
    \label{table-num_solutions}
    \multirow{1}{*}{\textbf{Method}} &
    \textbf{\BA} &
    \textbf{\brain} &
    \textbf{\hprd} &
    \textbf{\cosmos} &
    \textbf{\dblpS} &
    \textbf{\dblpL} & 
    \textbf{\ethe} & 
    \textbf{\youtubeS} & 
    \textbf{\youtubeL} \\
      \hline
    \ldf         & 1.21	& 2.05	& 0.69	& 4.80	& 4.82	& 1.51 & 2.50 & 2.73 & 1.75 \\
    \nlf       & 1.29	& 2.24	& 0.37	& 7.05	& 4.26	& 1.06 & 2.34 & 2.52 & 1.06 \\
    \gql        & 1.42	& 3.95	& 1.24	& \textbf{7.70}	& 5.23	& 2.55 & \textbf{3.53} & 2.73 & 1.23 \\
    \tso          & 1.41	& 1.26	& 0.50	& 5.53	& 4.36	& 1.36 & 2.30 & 2.57 & 1.24 \\
    \cfl & 1.56	& 1.12	& 0.18	& 6.97	& 5.70	& 1.13 & 2.32 & 2.84 & 0.90 \\
    \daf        & 1.31   & 2.11 & 0.54 & 6.93 & 4.86 & 2.63 & 2.58 & 3.32 & 1.88\\
    \ceci        & 1.21   & 1.03 & 1.05 & 6.96 & 4.18 & 1.35 & 1.70 & 1.60 & 0.55 \\
    \nmatch        & 1.02   & 0.00 & 0.00 & 0.24 & 1.25 & 0.24 & 1.39 & 0.37 & 0.00 \\

    \sgmrlmodel   & \textbf{2.37}	& \textbf{4.58}	& \textbf{4.17}	& 3.70	& \textbf{8.32}	& \textbf{6.24} & 2.53 & \textbf{7.93} & \textbf{3.62} \\
    \end{tabular}
  \end{center}
\vspace*{-6mm}
\end{table}

Given the nature of \task, we also evaluate the ability of each method to find as many solutions as possible. As Table~\ref{table-num_solutions} indicates, \sgmrlmodel outperforms baseline methods on 7 out of 9 datasets, suggesting that the slower 
but more intelligent \sgmrlmodel can not only ensure one solution is found, but also provide as many as and even more solutions compared against the faster baseline solvers. The exceptions are \hprd and \ethe, where \gql solves less pairs as shown in Figure~\ref{fig:result-curves}, but on average finds more solutions per graph pair across these two datasets. We empirically observe that the baseline solvers tend to find many solutions because as soon as one solution is found, nearby solutions can be found. In contrast, \sgmrlmodel continues performing neural network operations after one solution is found. This indicates the efficiency limitation of the current model, and calls for the need for future efforts to speed up the neural network computation, design even better policy network and training methods to allow discovery of more solutions in lesser iterations, etc.
\section{Conclusion}
\label{sec-conc}


In this paper, we tackle the challenging and important task of \task, and present a new method for efficient and effective exact \task. It contains a novel encoder-decoder neural network architecture trained using a novel look-ahead loss function and a max margin loss to improve the policy estimation as well as the encoder. The core component, \encoderlayer, is a query-conditioned graph encoder that performs intra-graph propagation and inter-graph node matching to capture each state. To address the reward sparsity issue posed by the large action space, we augment the positive node-node pairs at each search step with future states' node-node pairs. We experimentally show the utility of the proposed \sgmrlmodel method on the important \task task. Specifically, \sgmrlmodel is able to solve more graph pairs than several existing \task solvers on one synthetic dataset and \numdata large real-world datasets. 

\bibliographystyle{iclr2023_conference}

\bibliography{bibliography}

\appendix
\section{Details on Training and Testing of \sgmrlmodel}
\label{sec-train}

\subsection{Details on Collecting Training Signals}
\label{subsec-tr-signals}

As mentioned in the main text, we utilize the max margin loss proposed in \citet{lou2020neural} which can be written as
$$ L_{\mathrm{mm}}(s_{t+k}) = \sum_{(u, v) \in P({s_{t+k}})}   E(\bm{h}_u, \bm{h}_v) + \sum_{(u, v) \in N({s_{t+k}})}  \mathrm{max}\{0, \alpha - E(\bm{h}_u, \bm{h}_v)\} $$ where
$$ E(\bm{h}_u, \bm{h}_v) = ||\mathrm{max} \{ 0, \bm{h}_u - \bm{h}_v \} ||_2^2.$$ The intuition behind the loss function is to encourage the encoder to produce node embeddings that capture the subgraph-graph relationship. Specifically, the loss encourages each dimension of $\bm{h}_u$ to be less than $\bm{h}_v$, and the error term $E$ measures the deviation of the current node embeddings from this desired property.

It is noteworthy that both $L_{\mathrm{la}}$ and $L_{\mathrm{mm}}$ rely on the positive and negative node-node pairs, which are obtained by performing search on $(q,G)$ pairs where $q$ is the randomly sampled training subgraph queries from $G$. Therefore, the overall \sgmrlmodel model does not need any pre-solved graph pairs for training. 

Here we describe more details on how we prepare the positive node-node pairs $P(s_t)$. During training, we set a time limit of 5 minutes for each $(q,G)$ pair, and at the end of the search process, we obtain a search tree where each node in the search tree corresponds to a state $s_t$. For a state $s_t$, we look ahead at all sequences of states leading to a solution starting from $s_t$, and compute the $L_{\mathrm{la}}$ loss according to Equation~\ref{eq-L1} for each sequence, and sum up the loss for all such sequences. To obtain $N(s_t)$, we randomly sample $|P(s_t)|$ node-node pairs from the action space $\mathcal{A}_{u_t}$, where $u_t$ is the query graph node selected by $\phi$ as described in Section~\ref{subsec-search}, and $\mathcal{A}_{u_t}$ is the local candidate nodes in $G$ to be selected by the neural network.

\subsection{Details on Testing Graph Pairs}

We conduct evaluation on the following testing graph datasets:
\begin{enumerate}
    \item The synthetic dataset \BA is unlabeled and generated using the Barabási–Albert model~\citep{barabasi1999emergence} with the parameter of attachment set to 3. We sample 50 query graphs each of 64 nodes as the query set. The sampling algorithm starts with a randomly selected initial node, and then performs random walk until a subgraph consisting of a pre-defined number of nodes is sampled. Each step in the random walk selects a node from the neighboring nodes of the current node to visit in the next step, where the node can be a new node or an already selected node, i.e. the random walk is allowed to backtrack. We define the raw sampling probability to be $1/p$ if the node is the selected subgraph, and $p$ otherwise. A small $p$ therefore encourages re-visitation and the final subgraph tends to be ``star``-like, while a large $p$ typically yields a more ``path``-like subgraph. Each of the 50 sampled subgraphs uses a different $p$ parameter by sweeping across the range from $0.001$ to $1000$. Specifically, the $i$-th sampled subgraph uses $p=0.001*\mathrm{exp}\big( \mathrm{log}(10^{6})/49 * (i-1) \big)$, i.e. the first subgraph uses $p=0.001$, the second uses $p=0.00133$, the third uses $p=0.00176$, etc. The 49th subgraph uses $p=754.312$, and the last uses $p=1000$. Without specific mentioning, the sampling of subgraphs for other datasets uses this algorithm.
    \item The \brain dataset is obtained from mouse retina connectome~\citep{helmstaedter2013connectomic} with node labels representing the coarse type determined from supplemental text~\citep{brainDatasets}. We sample 50 query graphs each of of 64 nodes. There are 5 types of node labels in total. 
    \item The \hprd target graph is a protein-protein interaction graph obtained from a recent survey paper on \task~\citep{sun2020memory} without node labels. Similar to \brain, we sample 50 64-node graphs as the query graphs. 
    \item The \cosmos graph represents the network of galaxies~\citep{coutinho2016network} without node labels, for which we sample 50 query graphs each of 32 nodes as queries. 
    \item The \dblp and \youtube target graphs are also used in the survey paper~\citep{sun2020memory}, but instead of sampling subgraphs as query graphs, we rely on ground-truth communities mined from an algorithm~\citep{yang2012defining} as query graphs. According to \citet{snapnets}, for \dblp, authors who published to a certain journal or conference form a community, and for \youtube, user-defined groups are considered as ground-truth communities. Since many ground-truth communities are provided by \citet{snapnets}, we sample a 50-graph small query set and a 50-graph larger query set for these two datasets. Specifically, \dblpS contains communities of 20-30 nodes, \dblpL contains communities of 30-60 nodes, \youtubeS contains communities of 30-40 nodes, and \youtubeL contains communities of 40-100 nodes.
    \item The \ethe dataset is obtained from a recent paper on financial anomoly detection~\citep{chen2022antibenford}. The graph represents 1-week Ethereum blockchain transactions where each edge represents a transaction with the amount. We run the released code provided by the paper to extract top anomalous subgraphs, and filtering subgraphs that are too small or too large, yielding 24 subgraphs ranging from 6 to 822 nodes as query graphs. Intuitively, these subgraphs are relatively dense and correlate with possibly malicious behavior. We remove the transaction amount from the edges, and run \task using the mined subgraphs as query graphs and the original transaction graph as the target graph. Although \sgmrlmodel can potentially be extended for clique or anomaly detection, we leave the investigation as future work.
\end{enumerate}

\subsection{Details on Training and Validation Graph Pairs}

For each target graph, we sample query graphs of various sizes to perform curriculum learning, as shown effective by earlier graph matching works~\citep{bai2021glsearch, lou2020neural}. We ensure the testing query graphs and training query graphs are different by using different random seeds and further running an isomorphism checker to ensure none of the testing query graphs are visible during training. The query graphs come in the following sizes: 8, 16, 24, 32, 48, 64, 96, and 128 nodes. It is noteworthy that the query graphs are obtained from provided ground-truth communities or mined subgraphs for \dblp, \ethe and \youtube for testing, but during training, all the query graphs are obtained via subgraph sampling.

Regarding the validation scheme, we use the following validation set for validating the models: 3 query graphs of size 8, 3 query graphs of size 16, 3 query graphs of size 32, 3 query graphs of size 64, and 3 query graphs of size 128. These 15 query graphs are sampled from the same target graph with different random seeds, and we ensure they are not isomorphic to any query graphs in the training set (nor isomorphic to any query in the testing set if the testing set is obtained via random subgraph sampling).

The overall training process consists of the following iterative steps: (1) Sample a query graph from $G$ and perform search with a limit of 5 minutes; (2) Collect training signals and add them to a replay buffer; (3) Perform training by sampling a batch of training examples from the replay buffer and running back-propagation; (4) Once every 5 times of performing search on training graph pairs, we iterate through the 15 graph pairs in the validation set with a time budget of 40 seconds, and record the validation reward, which is defined as the average best subgraph size. We accept the trained model if the validation reward improves, and revert to the previous best model otherwise by loading the best model's checkpoint. At the end of the training process, we evaluate the best model that achieves the best validation performance.


\subsection{Details on Hyperparameters}
\label{subsec-hyperparam}

For the encoder, we set the intra-graph $f_{\mathrm{agg}}$ and $f_{\mathrm{msg}}$ functions to \textsc{GraphSAGE}~\citep{hamilton2017inductive}, and set the inter-graph $f_{\mathrm{agg}}$ function to be $\mathrm{SUM}$ and $f_{\mathrm{msg}}$ function to be dot-product style attention~\citep{vaswani2017attention} over incoming messages. 
Specifically, the message collected by each query node, $\bm{h}_{u,G\rightarrow q} \in \mathbb{R}^{D}$, consists of attended node embeddings from $v'\in \tilde{M_t}(u)$. Overall, $\bm{h}_{u,G\rightarrow q}$ is computed as
\begin{equation}
\sum_{v' \in \tilde{M_t}(u)} \mathrm{softmax}_{v'\in \tilde{M_t}(u)}\big( \mathrm{MLP}_q(\bm{h}_{u,\mathrm{intra}})^T \mathrm{MLP}_G(\bm{h}_{v',\mathrm{intra}}) \big) \mathrm{MLP}_{\mathrm{VAL},q}(\bm{h}_{v',\mathrm{intra}}) .
\end{equation}
The message collected by each target node,
$\bm{h}_{v,q\rightarrow G}  \in \mathbb{R}^{2D}$, consists of attended node embeddings from $u' \in \tilde{M_t\inv}(v)$ and the whole query graph embedding, $\bm{h}_{q, \mathrm{intra}}$. Overall, $\bm{h}_{v,q\rightarrow G}$ is computed as 
\begin{equation}
\bm{h}_{v,q\rightarrow G} =\sum_{u' \in \tilde{M_t\inv}(v)}
\mathrm{CONCAT} (\bm{h}_{v,q\rightarrow G}', \bm{h}_{q, \mathrm{intra}}),
\end{equation}
where
\begin{equation}
\bm{h}_{v,q\rightarrow G}' =
\mathrm{softmax}_{u' \in \tilde{M_t\inv}(v)} \big(  \mathrm{MLP}_q(\bm{h}_{u',\mathrm{intra}})^T \mathrm{MLP}_G(\bm{h}_{v,\mathrm{intra}}) \big) \mathrm{MLP}_{\mathrm{VAL},G}(\bm{h}_{u',\mathrm{intra}}).
\end{equation}
In total, we use four different MLPs, i.e. $\mathrm{MLP}_q$, $\mathrm{MLP}_G$, $\mathrm{MLP}_{\mathrm{VAL},q}$, and $\mathrm{MLP}_{\mathrm{VAL},G}$ following \citep{vaswani2017attention}, with an $\mathrm{ELU}(\cdot)$ activation after the final layer.

We set the $f_{\mathrm{combine}}$ function to be the concatenation $\mathrm{CONCAT}$ followed by $\mathrm{MLP}$. We choose $\mathrm{MEAN}$ as the readout function $f_{\mathrm{readout}}$ for computing the query graph-level embedding. We choose the $\mathrm{MAX}$ function as the aggregation function for our Jumping Knowledge network which aggregates the node embeddings. We apply layer normalization~\citep{ba2016layer} to the final node embeddings before feeding into the decoder. 

For the encoder, we stack 8 layers of intra-graph message passing with dimension 16. We stack 8layers of the inter-graph message passing with all $\mathrm{MLP}$s outputting dimension 16. For the decoder, we set $\mathrm{MLP}_{\mathrm{att}}$ to dimensions [16, 4, 1] and $\mathrm{MLP}_v$ to dimensions [16, 32, 16, 8, 4, 1]. We set the policy head's bilinear layer, $\bm{W}^{[1:F]}$, to output dimension $F=32$ and the policy head's $\mathrm{MLP}$ to dimensions [48, 32, 16, 8, 1]. For all the $\mathrm{MLP}$s, we use the $\mathrm{ELU}(x)$ activation function on the hidden layers. For $\mathrm{MLP}_v$, we apply a $\mathrm{LeakyReLU}(\cdot)$ function following the final layer.

The goal of the encoder is to learn good node embeddings that capture the structural matching of $q$ and $G$, and thus we do not encode the node labels as initial encodings. Instead, we rely on the candidate set generation algorithm to handle the constraint that only nodes of the same label can match. Therefore, it is critical to encode the nodes into initial encodings properly. A simple way to encode the nodes is to assign a constant encoding to every node in both $q$ and $G$. To enhance the initial encodings, we adopt Local Degree Profile (LDP) scheme~\citep{cai2018simple} concatenated with a 2-dimensional one-hot vector indicating whether a node is currently selected or not.

Our replay buffer is of size 128. During the training process, we repeatedly run 5 minutes of search, populating the replay buffer, 3 minutes of replay buffer sampling and back-propagation, training on as many replay buffer samples as possible, and performing validation to choose whether or not we commit the learned model. One iteration of this process typically takes around 10 minutes. Hyperparameters were found by tuning for performance manually against the validation dataset.




\section{Notes on Search}
\label{sec-search}


\subsection{Induced vs Non-Induced Subgraphs}

\begin{wrapfigure}{r}{0.4\textwidth}
\vspace{-1em}
\centering
  \includegraphics[width=0.4\textwidth]{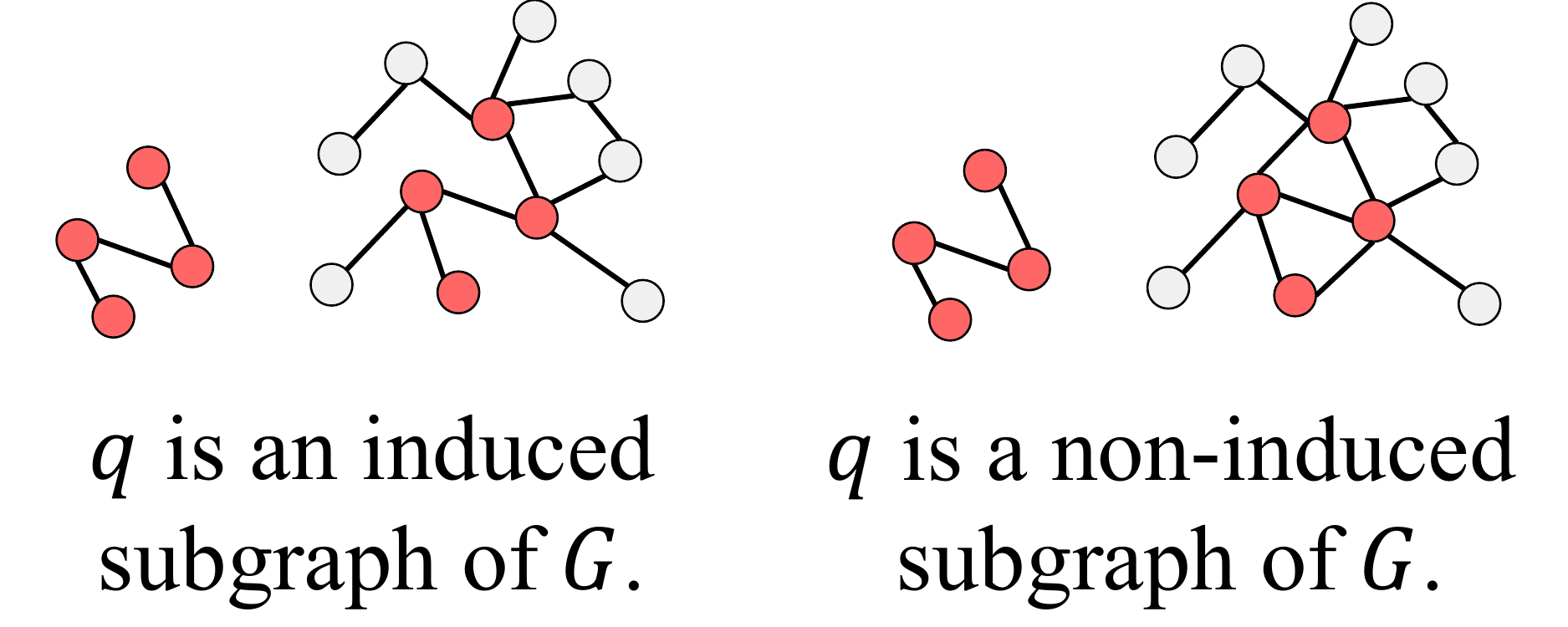}
\caption{Comparison between an induced (left) and a non-induced (right) subgraph. We use the red color to denote the nodes that are included in the match.}
\label{fig:induced}
\end{wrapfigure}
One key detail regarding the subgraph definition is whether the subgraph is induced or non-induced.
An induced subgraph $q[S]$ consists of all the nodes $S \subseteq V_q$ and requies that for every edge $(u,u') \in E_q$, if both $u$ and $u'$ are in $S$, then the edge $(u,u')$ must be included in the edges of $q[S]$ as well. Intuitively, an induced subgraph does not allow any edges in the original graph to be dropped, as long as both endpoints of the edge are included by the subgraph. In contrast, a non-induced subgraph $q[S]$ consists of all the nodes $S \subseteq V_q$ but allows edges in $E_q$ to be dropped. In this paper, we adopt the definition of non-induced subgraph, which is consistent with \citet{han2019efficient}. 

\subsection{Choice of Filtering, Query Node Ordering, and Local Candidate Computation}

Motivated by prior work benchmarking solver baselines~\citep{sun2020memory}, we adopt \daf as the filtering method, \gql as the query node ordering method, and \gql as the local candidate computation method in \sgmrlmodel and \nmatch. As shown by the experimental results in the main paper, this design choice plays a much smaller role in \task than smart target graph node ordering, highlighting the need for \sgmrlmodel. Furthermore, we find that optimal filtering, query node ordering, and local candidate computation settings are highly dataset-dependent and hard to determine beforehand; however, because \sgmrlmodel's design is orthogonal to such choices, it could easily be adapted to different filtering and query node ordering algorithms. We leave such study to future work. 


The filtering method produces a global candidate set, $C: V_q \rightarrow V_t$, to drastically prune the search space of possible matchings. A simple method to compute $C$ is to only allow mappings from $u_t$ to $v_t$ if $v_t$ has degree higher than or equal to the degree of $u_t$. Different filtering algorithms use different such rules~\citep{sun2020memory}. The query node ordering gives a search plan on which $u$ nodes to select first. A simple method is picking the $u$ nodes with the least candidate target nodes given by $C$. Different node ordering algorithms use different such rules~\citep{sun2020memory}.

During the search process, the local candidate set guarantees all subgraph mappings found throughout the search process are isomorphic. Specifically, it ensures $u_t$ is matchable to $v_t$, i.e. $v_t \in \mathcal{A}_{u_t}$ only if (1) all edges between $u_t$ and the currently matched query subgraph (with node set $S$) exist between $v_t$ and the currently matched target subgraph: $(u_t, u')\in E_q, u' \in S \implies (v_t,M(u'))\in E_G$ and (2) the matching, $(u_t, v_t)$ exists in the candidate set: $v_t \in C(u_t)$. There are many different implementations and optimizations to ensure these conditions hold. For instance, a query tree data structure~\citep{han2019efficient} can be stored, allowing $O(1)$ lookup for $(u_t, u')\in E_q$ connections given a particular query node ordering. Different local candidate computation algorithms use different such optimizations~\citep{sun2020memory}. Even though the subgraph isomorphism constraint is ensured through the local candidate set, \task is by no means an easy task due to the large amount of nodes in $\mathcal{A}_{u_t}$.

Among different techniques to further improve the accuracy of \task approaches, learning a state-dependent policy to select nodes from $q$ is one direction worth mentioning. We currently only learn to select nodes in $\mathcal{A}_{u_t} \subseteq V_G$ partly because there is usually a small amount of nodes in $q$, e.g. 32 nodes or 128 nodes, while the number of nodes in $\mathcal{A}_{u_t}$ is much larger.
As mentioned in the main text, RL-QVO~\citep{wang2022reinforcement} learns a state-independent node ordering policy, $\phi$, via RL that is executed before search. We call such $\phi$ query node ordering. We argue it is a promising future direction to explore learning to select both nodes in $q$ and nodes in $G$ at each search iteration.

As shown effective by previous works, we adapt a promise-based search strategy~\citep{bai2021glsearch}, which backtracks to any earlier search state, instead of the immediate parent, whenever a terminal state is reached. We choose which earlier search state to backtrack to by computing a 2:1 weighted average of the search depth normalized by the query graph size and the percentage of explored actions, allowing search to quickly exit local minima. We include this adaptation on all learning methods. We also tried the promise-based search on the solver baselines, but found the performance difference is minimal, typically around 2\%. This is because solver baselines do not assign any ordering to the target graph nodes, thus there is no clear incentive to backtracking early. 

\section{Comparison with Maximum Common Subgraph (MCS) Detection}
\label{sec-related}


\task and MCS detection are highly related tasks, and one can convert \task to MCS detection by feeding the input $(q,G)$ pair to an MCS solver, and check if the MCS between $q$ and $G$ is identical to $q$. However, we note the following differences at the task level:
\begin{enumerate}
    \item \task requires all the nodes and edges in $q$ to be matched with $G$, whereas MCS detection does not require one of two input graphs to be contained in another.
    \item \task by definition requires $q$ to be smaller or equal to $G$, since otherwise the solver can immediately return ``no solution''. In practice, the query graph is usually given by the user as input, and usually contains less than 100 nodes~\citep{sun2020memory}. 
\end{enumerate}

The first difference has several consequences. First, although one can solve \task via an MCS solver, the other way does not hold, i.e. one cannot use a \task solver for MCS detection, since \task has the stronger constraint of matching the entire $q$. Second, the stronger constraint of \task allows existing search algorithms to design pruning techniques as mentioned in the main text.

The second difference implies that the learning models designed for \task may not work well for MCS detection and vice versa. For example, \sgmrlmodel has a novel encoder layer with a matching module that has time complexity $\mathcal{O}\big(|V_q||\bar{\mathcal{A}}_{u_t}|\big)$. However, if $q$ becomes prohibitively large, e.g. as large as $G$ containing millions of nodes, then the matching step would become a bottleneck and make the overall approach too slow to be useful in practice. Therefore, \sgmrlmodel cannot be used for MCS detection not only because fundamentally the search algorithm leverages the stronger constraint if \task, but also because the learning model would be inefficient and thus useless in practice. By the same reasoning, \sgmrlmodel is likely to be ineffective for graph isomorphism checking when both input graphs are of the same size and very large. This suggests the difficulty of designing a general neural network architecture that works efficiently and effectively for a series of related but different combinatorial optimization tasks. In theory, many such tasks are equivalent and can be converted to each other, but in practice, specialized models should be designed to render a truly useful learning-based approach.

A similar argument can be made for \mcsrlmodel~\citep{bai2021glsearch} designed for MCS detection. \mcsrlmodel treats both input graphs symmetrically and does not leverage the fact that all the nodes and edges in $q$ must be matched in $G$. The learning component \mcsrlmodel assumes both input graphs can be very large, and therefore instead of outputting a policy \policye for every action $a_t$ at $s_t$, \mcsrlmodel performs the execution of each action to get its next state $s_{t+1}$, and compute the value associated with $s_{t+1}$ for the $Q(s_t,a_t)$. This is named as ``factoring out action'' in \citet{bai2021glsearch}, and requires sequentially going through all the actions to obtain the next states, for which \mcsrlmodel adopts a heuristic to reduce the amount of actions. To adapt \mcsrlmodel for \task, one has to use a heuristic to reduce the local candidate space, i.e. the action space, in order to be efficient enough to compare against existing solvers. In contrast, \sgmrlmodel efficiently computes one embedding per node for both $q$ and $G$ using the \encoderlayer encoder layer, and computes \policye in the decoder by batching the node embeddings that are in the action space, i.e. $P_{\mathrm{logit}}(a_t|s_t)=\mathrm{MLP} \big( \mathrm{CONCAT} (\bm{h}_{u_t}^{T} \bm{W}^{[1:F]} \bm{h}_{v_t}, \bm{h}_{s_t} ) \big)$ can be turned into a batch operation for all the node embeddings involved in the action space of $s_t$. The \valuee is only needed during the training stage to enhance \policye, and thus does not need to be predicted during inference.

In short, \sgmrlmodel cannot be used for large-input MCS detection due to the matching module performing matching between nodes in $q$ and $G$, while \mcsrlmodel may be adapted for \task, it is likely to be ineffective due to the lack of policy network to efficiently compute a score for each pair of nodes leveraging the constraint of \task. 

\section{Ablation Studies}
\label{sec-ablation}

We perform a series of ablation studies whose results are shown in Table~\ref{table-ablation}. We find our key novelties indeed greatly contribute to \sgmrlmodel's superior performance on \task, particularly the encoder-decoder design.

\begin{table}[h]
  \begin{center}
    \caption{Ablation study results on \task over \hprd. The ratio of the number of solved pairs over the 50 pairs in the query set is reported. The average number of solutions (divided by $10^3$) is reported too.
    }

    \begin{tabular}{l|p{2cm}p{2cm}}
    \label{table-ablation}
    \textbf{Model} & \textbf{Solved \%} & \textbf{\# Solutions} \\
    \hline
    
    \ldf & 0.08 & 0.69 \\ 
    \nlf & 0.04 & 0.37 \\ 
    \gql & 0.16 & 1.24 \\ 
    \tso & 0.06 & 0.50 \\ 
    \cfl & 0.02 & 0.18 \\ 
    \daf & 0.06 & 0.54 \\ 
    \ceci & 0.12 & 1.05 \\ 
    \nmatch & 0.00 & 0.00 \\ 
    \sgmrlmodel & \textbf{0.48} & \textbf{4.17} \\ \hline
    
    \sgmrlmodel-rand-neg & 0.36 & 3.35 \\ 
    \sgmrlmodel-no-look-ahead & 0.42 & 3.68 \\ 
    \sgmrlmodel-no-mm & 0.08 & 0.73 \\ 
    \sgmrlmodel-2-encoder-layers & 0.24 & 0.25 \\ 
    \sgmrlmodel-$D=4$ & 0.04 & 0.30 \\ 
    \sgmrlmodel-\textsc{Gin} & 0.30 & 2.73 \\ 
    \sgmrlmodel-no-Q-matching & 0.34 & 2.98 \\
    \sgmrlmodel-no-LDP-encode & 0.26 & 2.39 \\
    
    \end{tabular}
  \end{center}
\end{table}

Table~\ref{table-ablation} shows the ablation study results. 
\begin{enumerate}
    \item We first perform the random sampling of negative node-node pairs ($N(s_t)$ in Equation~\ref{eq-L1prime}) from all the node-node pairs in $V_q$ and $V_G$ instead of from $\mathcal{A}_{u_t}$. We denote the model as ``\sgmrlmodel-rand-neg'', which performs worse than \sgmrlmodel but still outperforms baselines. This suggests that the negative sampling should be performed using ``harder'' examples sampled from the local candidate space instead of all the possible node-node pairs. Such harder negative examples train the model to better identify the promising actions over the unpromising ones at each state.
    \item We replace the look-ahead loss with a simpler version, by only looking at the current state's positive node-node pairs without considering future states. \sgmrlmodel-no-look-ahead performs worse than \sgmrlmodel, indicating the usefulness of the look-ahead loss.
    \item We remove the max margin loss, and as a result, \sgmrlmodel-no-mm performs much worse, suggesting the necessity for training the encoder to produce good node embeddings. Since the encoder training is shown to be important, we then perform a series of ablation study experiments on the encoder.
    \item \sgmrlmodel-2-encoder-layers uses 2 encoder layers instead of 8. \sgmrlmodel-$D=4$ decreases the embedding dimension from 8 to 4. \sgmrlmodel replaces the $\textsc{GraphSAGE}$ model with $\textsc{Gin}$. These 3 models perform worse than \sgmrlmodel, among which \sgmrlmodel-$D=4$ performs the worst, indicating that a dimension that is too small cannot capture enough information for the \hprd dataset.
    \item \sgmrlmodel-no-Q-matching removes the concatenation of $\bm{h}_{q,\mathrm{intra}}^{(k+1)}$ in the computation of $\bm{h}_{v,q \rightarrow G}^{(k+1)}$ in Equation~\ref{eq-encoderlayer}. The worse performance shows the contribution of making the attention from $q$ to $G$ conditioned on the whole-graph representation of $q$. 
    \item \sgmrlmodel-no-LDP-encode removes the Local Degree Profile encoding mentioned in Section~\ref{subsec-hyperparam}. The worse performance shows the importance of encoding the nodes initially with degree information.
\end{enumerate}

\section{Result Visualization}

\begin{figure}[h]
     \centering
    \begin{tabular}{c|c}
     \includegraphics[width=0.3\textwidth]{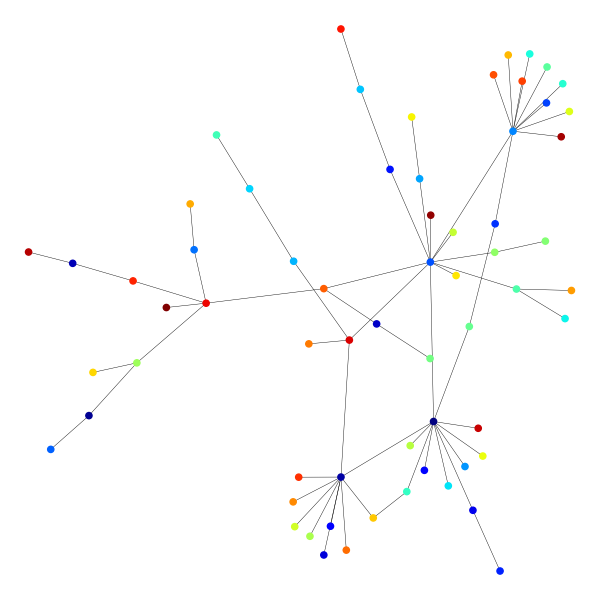} &
     \includegraphics[width=0.7\textwidth]{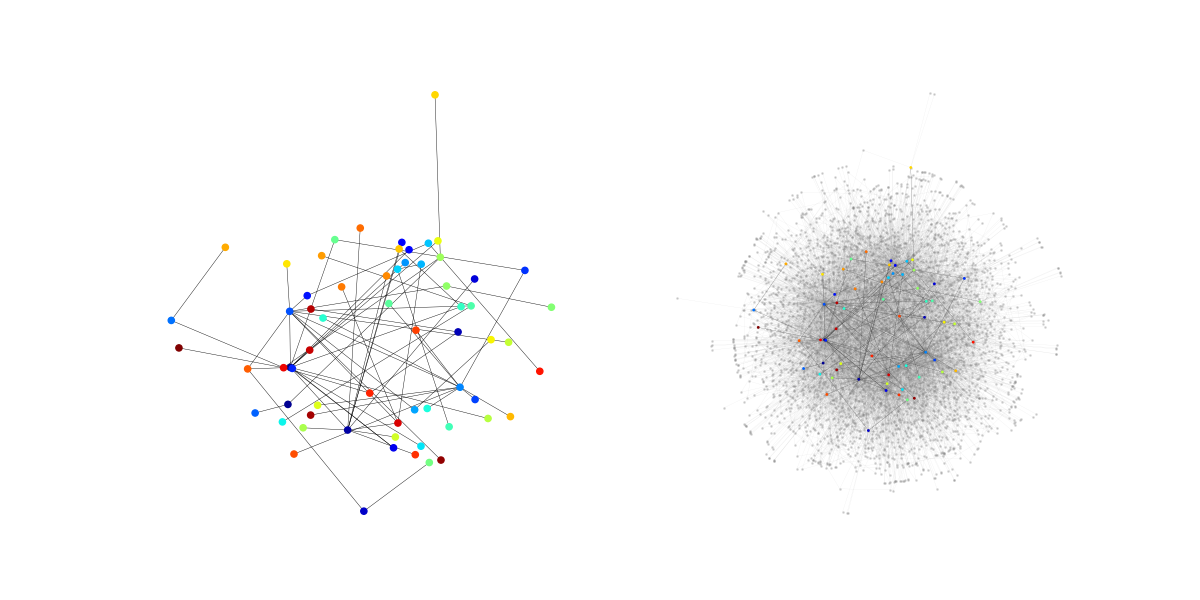}
     \end{tabular}
     \caption{Visualization of a solved pair on \BA. }
     \label{fig:vissol_ba}
\end{figure}

\begin{figure}[h]
     \centering
    \begin{tabular}{c|c}
     \includegraphics[width=0.3\textwidth]{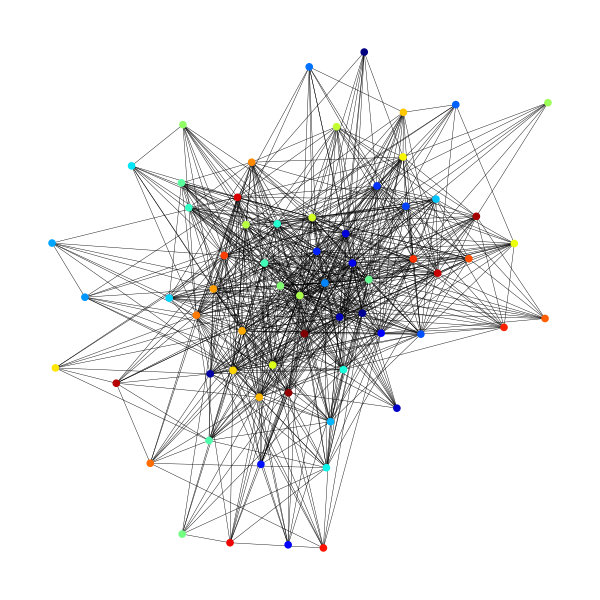} &
     \includegraphics[width=0.7\textwidth]{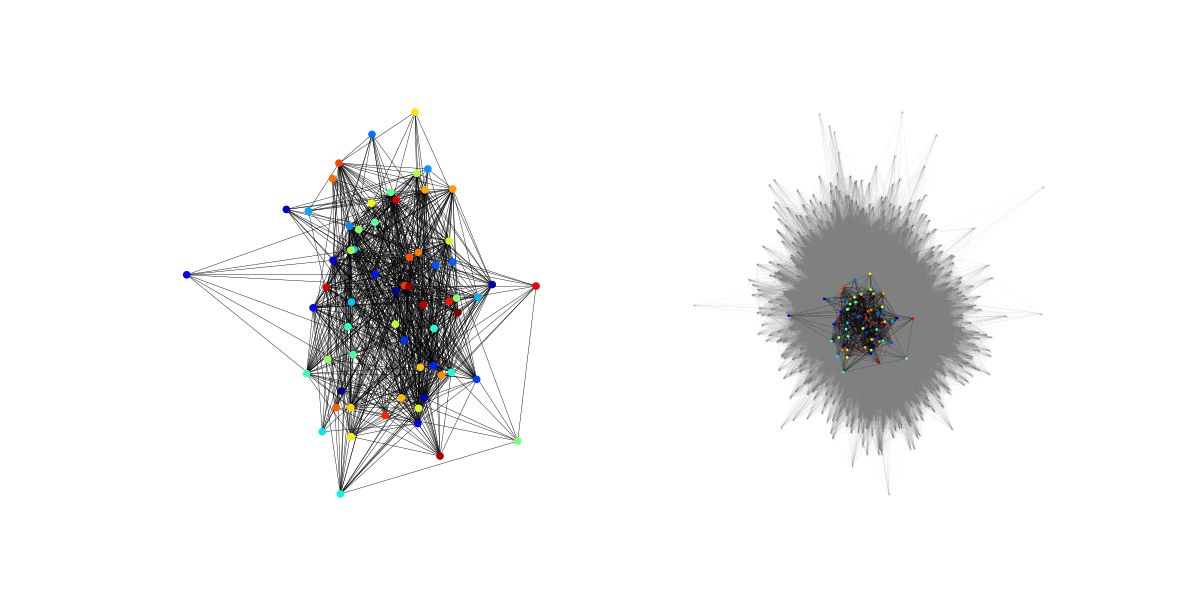}
     \end{tabular}
     \caption{Visualization of a solved pair on \brain. }
     \label{fig:vissol_brain}
\end{figure}

\begin{figure}[h]
     \centering
    \begin{tabular}{c|c}
     \includegraphics[width=0.3\textwidth]{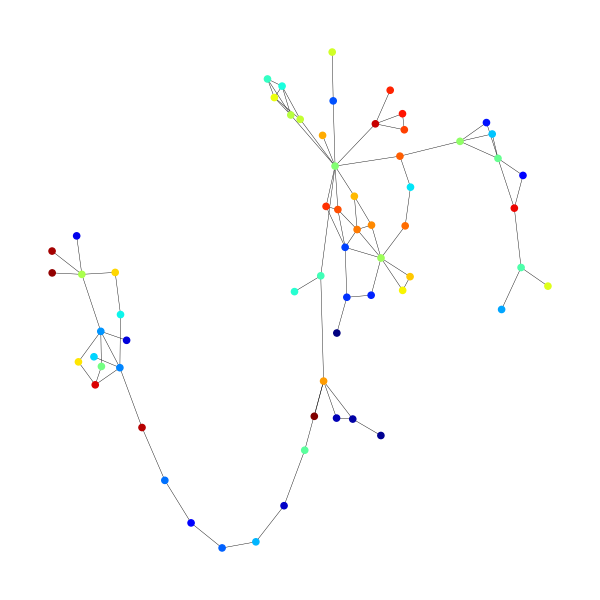} &
     \includegraphics[width=0.7\textwidth]{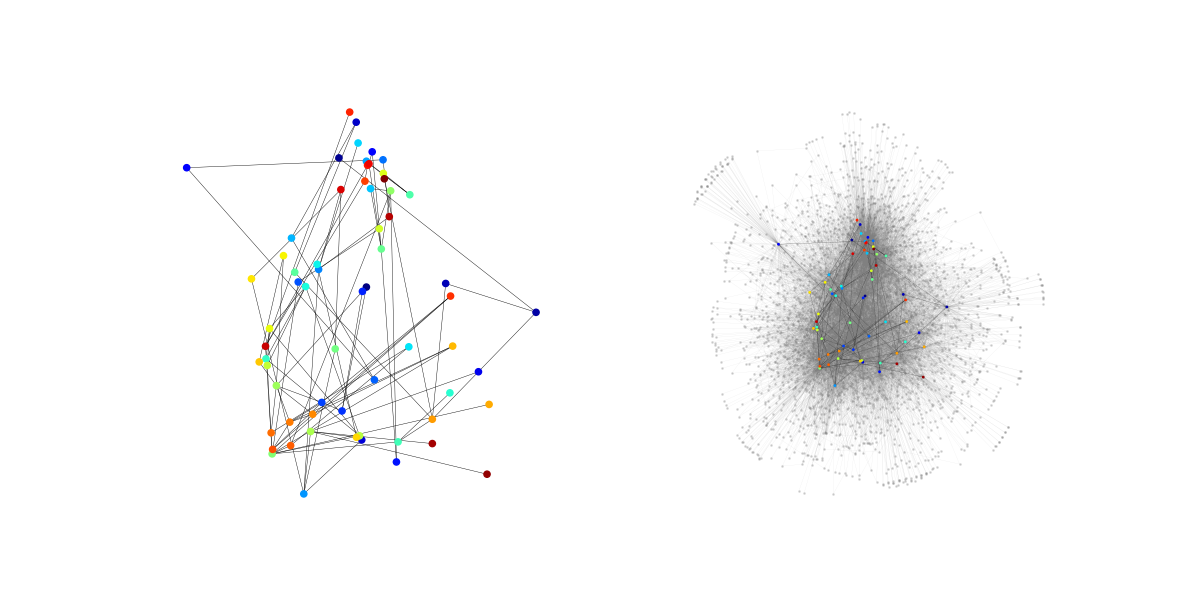}
     \end{tabular}
     \caption{Visualization of a solved pair on \hprd. }
     \label{fig:vissol_hprd}
\end{figure}

\begin{figure}[h]
     \centering
    \begin{tabular}{c|c}
     \includegraphics[width=0.3\textwidth]{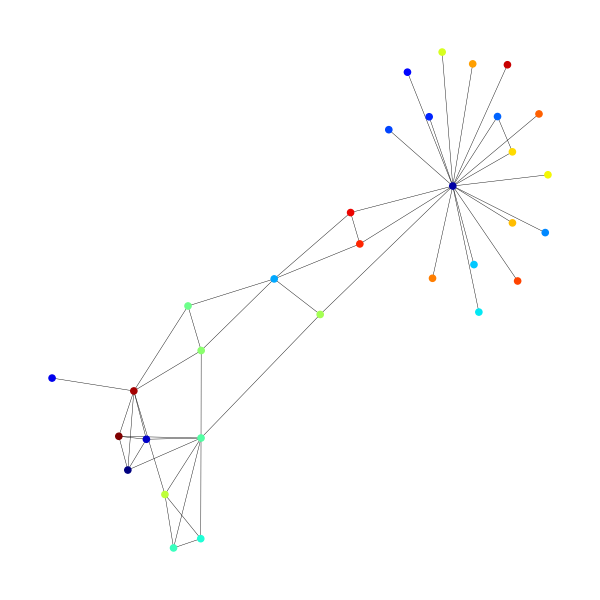} &
     \includegraphics[width=0.7\textwidth]{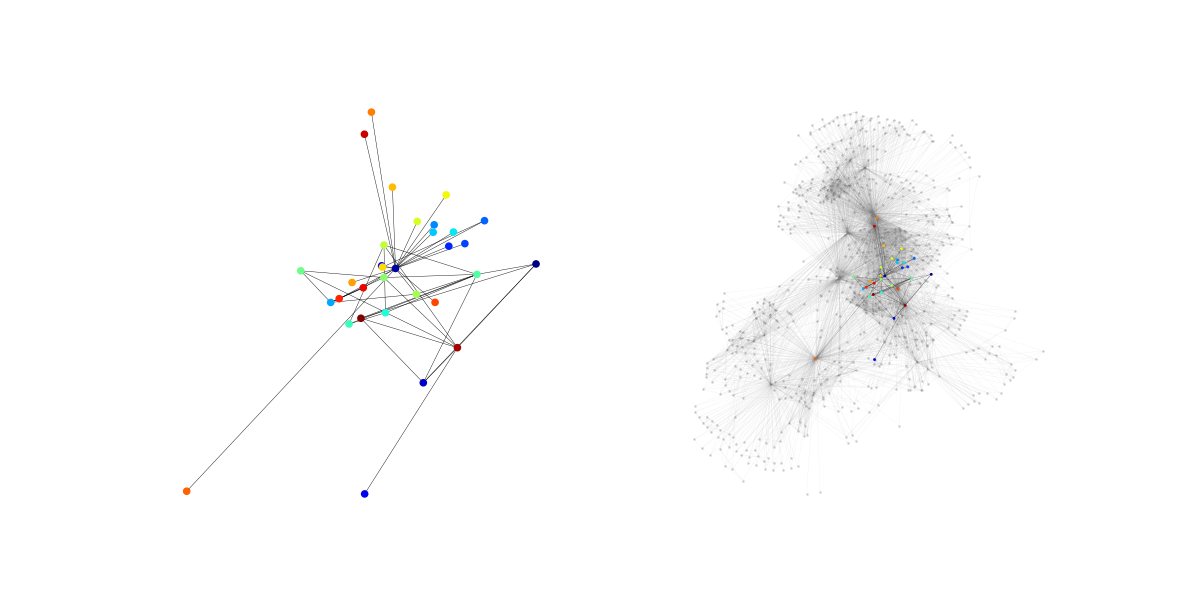}
     \end{tabular}
     \caption{Visualization of a solved pair on \cosmos. }
     \label{fig:vissol_cosmos}
\end{figure}

\begin{figure}[h]
     \centering
    \begin{tabular}{c|c}
     \includegraphics[width=0.3\textwidth]{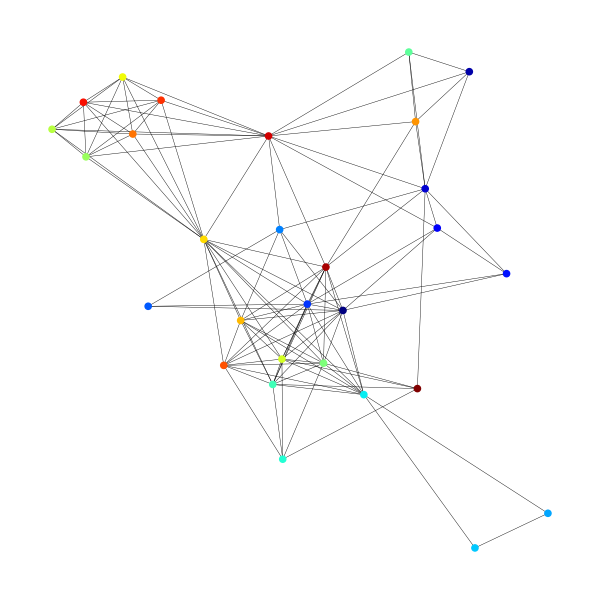} &
     \includegraphics[width=0.7\textwidth]{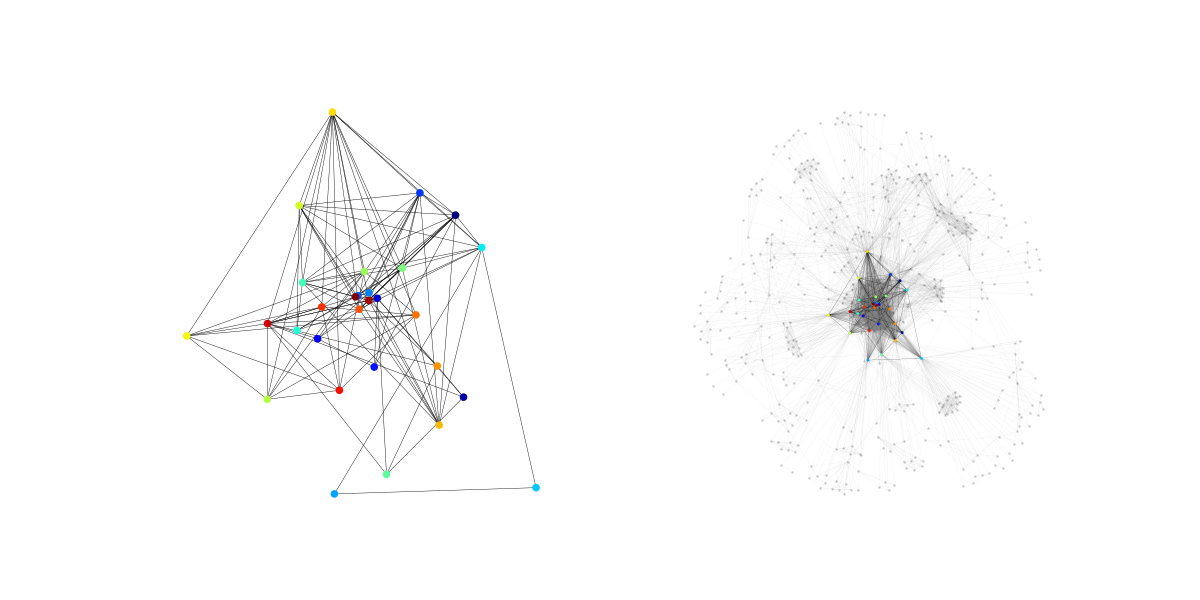}
     \end{tabular}
     \caption{Visualization of a solved pair on \dblp. }
     \label{fig:vissol_dblp}
\end{figure}

\begin{figure}[h]
     \centering
    \begin{tabular}{c|c}
     \includegraphics[width=0.3\textwidth]{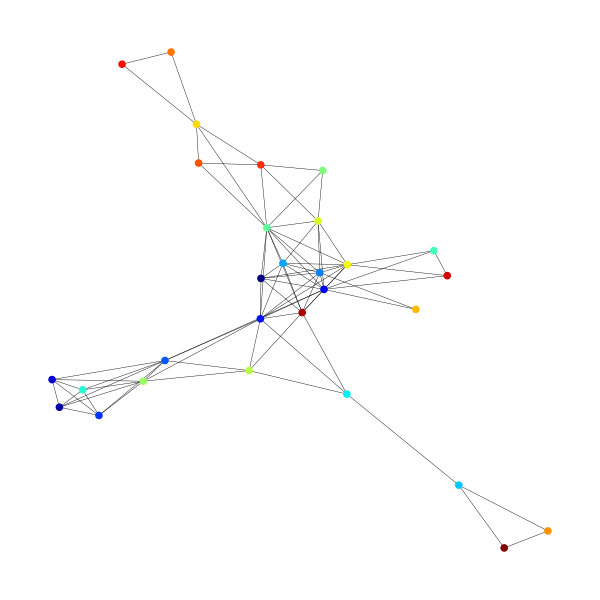} &
     \includegraphics[width=0.7\textwidth]{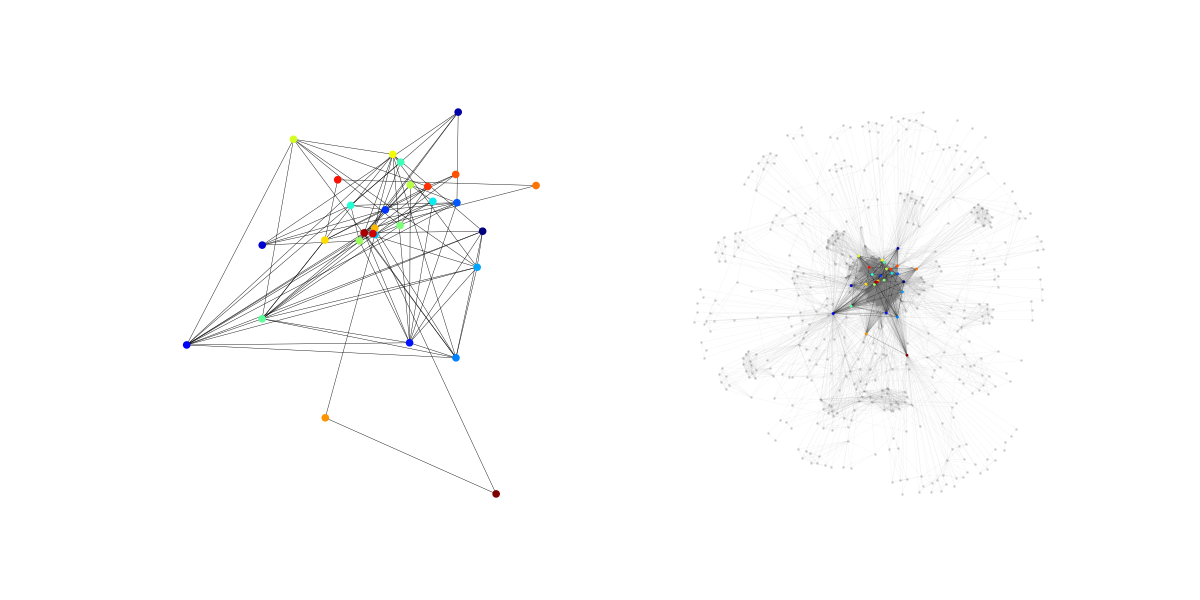}
     \end{tabular}
     \caption{Visualization of a solved pair on \dblp. }
     \label{fig:vissol_dblp2}
\end{figure}

\begin{figure}[h]
     \centering
    \begin{tabular}{c|c}
     \includegraphics[width=0.3\textwidth]{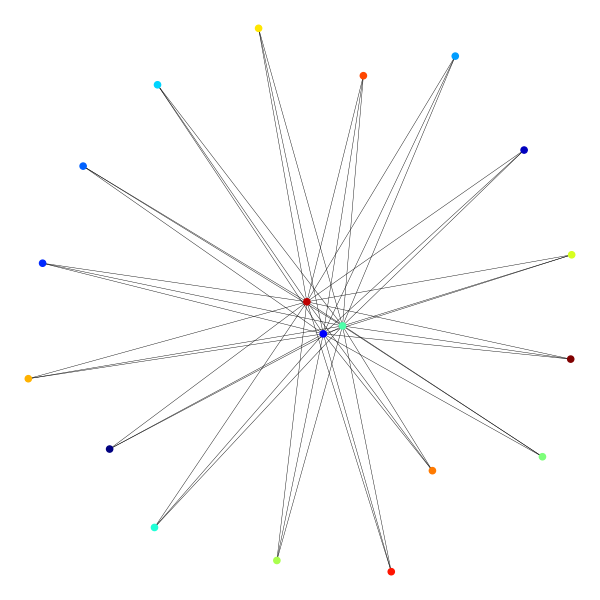} &
     \includegraphics[width=0.7\textwidth]{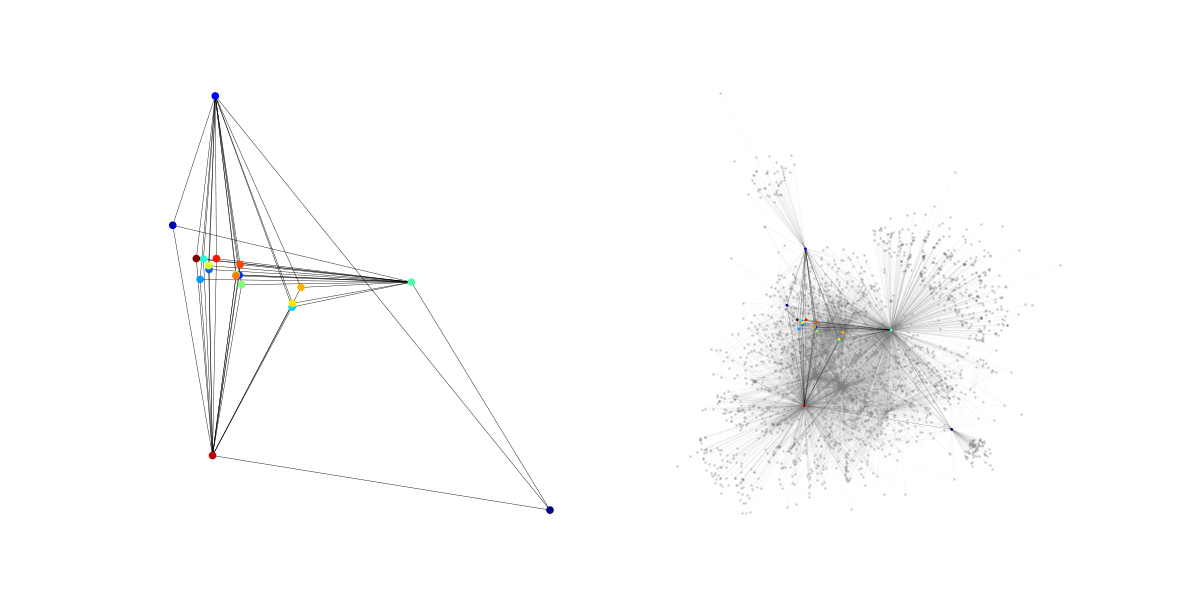}
     \end{tabular}
     \caption{Visualization of a solved pair on \ethe. }
     \label{fig:vissol_ethe}
\end{figure}

\begin{figure}[h]
     \centering
    \begin{tabular}{c|c}
     \includegraphics[width=0.3\textwidth]{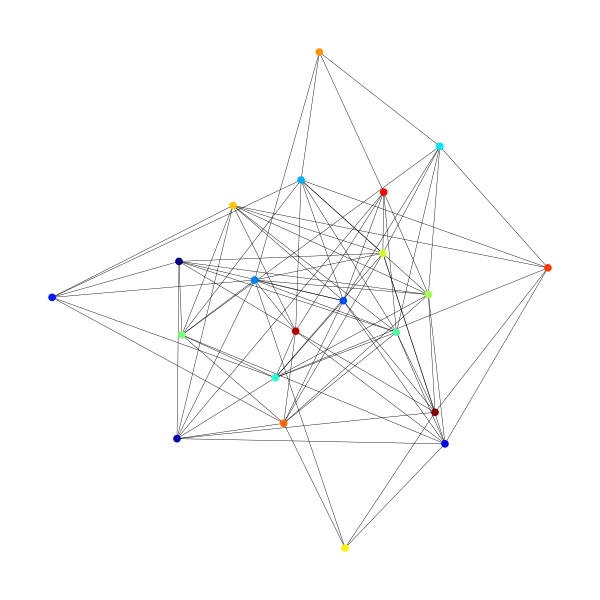} &
     \includegraphics[width=0.7\textwidth]{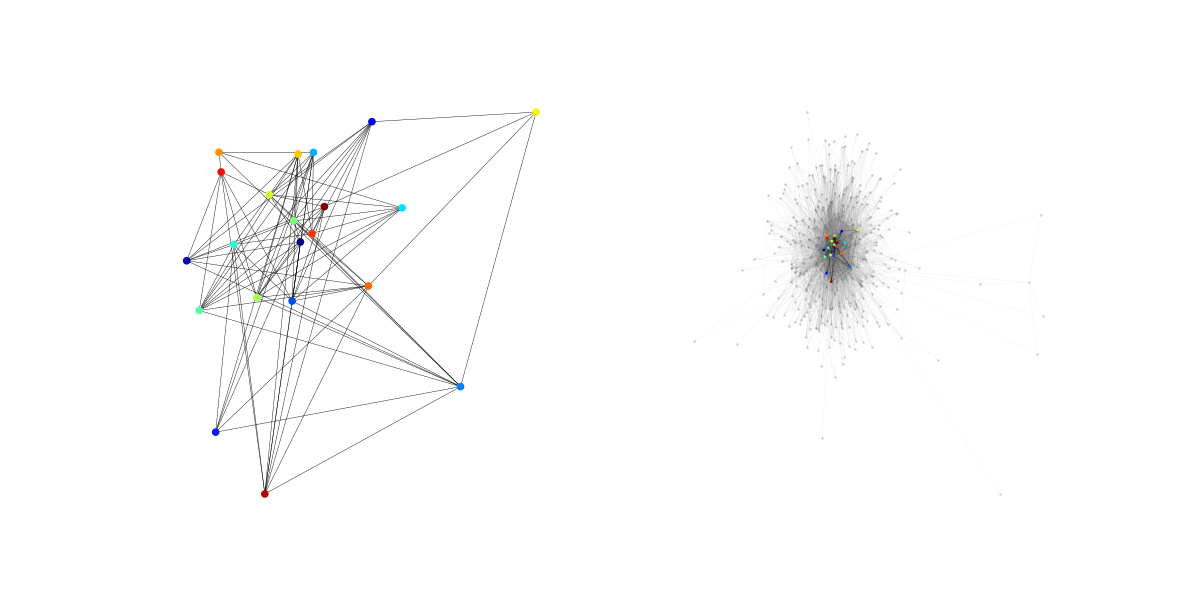}
     \end{tabular}
     \caption{Visualization of a solved pair on \ethe. }
     \label{fig:vissol_ethe2}
\end{figure}

We show five instances of solved graph pairs by \sgmrlmodel in Figures~\ref{fig:vissol_ba}, \ref{fig:vissol_brain}, \ref{fig:vissol_hprd}, \ref{fig:vissol_cosmos}, \ref{fig:vissol_dblp}, \ref{fig:vissol_dblp2}, \ref{fig:vissol_ethe} and \ref{fig:vissol_ethe2}. We first plot (1) the query graph on its own, then plot (2) the query graph matched to the target graph, and finally plot (3) the target graph.
In other words, the first and second plots correspond to the same query graph but their node positions/layouts are different due to different layouts. In the second plot we fix the positions of nodes in $q$ to match the positions of their matched nodes in $G$, i.e. the nodes in the second and the third plots have the same relative node positions for visualizing the node-node mapping. The colors of nodes are for the purpose of visualizing the mapping. For clarity, we only show a subgraph of $G$ instead of the entire $G$ as it contains too many nodes and edges to show. Specifically, we include the matched $q$ in $G$ and grow the matched subgraph by including the first-order neighbors of the matched nodes. 

\clearpage

\end{document}